\pdfoutput=1
\documentclass[11pt]{article}

\usepackage[]{emnlp2023}
\usepackage{times}
\usepackage{latexsym}
\usepackage{amsmath}
\usepackage{amssymb}
\usepackage{enumitem}
\usepackage{algorithm}
\usepackage{algpseudocode}
\usepackage[T1]{fontenc}
\usepackage[utf8]{inputenc}

\usepackage{microtype}

\usepackage{inconsolata}

\usepackage{comment}
\usepackage{graphicx}
\usepackage{multirow}
\usepackage{hyperref}
\usepackage{subcaption}
\usepackage{multicol}
\makeatother

\usepackage{arydshln}
\usepackage{xcolor,colortbl}

\usepackage{tabularx}

\usepackage{array}

\usepackage{makecell}

\definecolor{Gray}{gray}{0.90}
\newcolumntype{a}{>{\columncolor{Gray}}c}

\newcommand{\ours}{\textsc{InFusE}}
\newcommand{\sentli}{\textsc{SeNtLI}}
\newcommand{\oursub}{\textsc{InFusE}$_\textsc{sub}$}

\newcommand{\fulldoc}{\textsc{FullDoc}}
\newcommand{\summac}{\textsc{Summac}}

\makeatletter
\newcommand{\thickhline}{%
    \noalign {\ifnum 0=`}\fi \hrule height 1pt
    \futurelet \reserved@a \@xhline
}
\makeatother

\makeatletter
\def\Cline#1#2{\@Cline#1#2\@nil}
\def\@Cline#1-#2#3\@nil{%
  \omit
  \@multicnt#1%
  \advance\@multispan\m@ne
  \ifnum\@multicnt=\@ne\@firstofone{&\omit}\fi
  \@multicnt#2%
  \advance\@multicnt-#1%
  \advance\@multispan\@ne
  \leaders\hrule\@height#3\hfill
  \cr}
\makeatother

\title{Fine-Grained Natural Language Inference Based\\ Faithfulness Evaluation for Diverse Summarisation Tasks}

\author{Huajian Zhang\thanks{~~Part of the work done for his MSc thesis at the University of Edinburgh.} \quad Yumo Xu\thanks{~~Work done while at the University of Edinburgh.} \quad Laura Perez-Beltrachini \\ ILCC, School of Informatics \\
   University of Edinburgh\\
   \texttt{huajian.zhang.21@gmail.com}, \texttt{\{yumo.xu,lperez\}@ed.ac.uk}}
   
\begin{document}
\maketitle
\begin{abstract}
We study existing approaches to leverage off-the-shelf Natural Language Inference (NLI) models for the evaluation of summary faithfulness and argue that these are sub-optimal due to the granularity level considered for premises and hypotheses. That is, the smaller content unit considered as hypothesis is a sentence and premises are made up of a fixed number of document sentences.
We propose a novel approach, namely \ours, that uses a variable premise size and simplifies summary sentences into shorter hypotheses. 
Departing from previous studies which focus on single short document summarisation, we analyse NLI based faithfulness evaluation for diverse summarisation tasks. We introduce DiverSumm, a new benchmark comprising long form summarisation (long documents and summaries) and diverse summarisation tasks (e.g., meeting and multi-document summarisation). 
In experiments, \ours~ obtains superior performance across the different summarisation tasks.
\footnote{
Our code and data are available at \url{https://github.com/HJZnlp/infuse}}

\end{abstract}

\section{Introduction}

Current state-of-the-art summarisation systems are able to generate fluent summaries; however, 
their inability to generate factually consistent summaries remains a significant constraint in their real-world applications.
As a result, the assessment of  \textit{summary faithfulness}, i.e., the degree to which a summary accurately represents the content of the input document, has recently received much research attention. This evaluation is key to assess progress in abstractive summarisation \cite{gehrmann-etal-2021-gem,10.1613/jair.1.13715}. Existing research focuses on developing models to detect unfaithful summary content (\citealt{kryscinski-etal-2020-evaluating,scialom-etal-2021-questeval,ribeiro-etal-2022-factgraph}; \textit{inter alia})
as well as the meta-evaluation of these models with better benchmarks \cite{chen-etal-2021-factuality-checkers,honovich-etal-2022-true,durmus-etal-2022-spurious}. 

One way increasingly adopted to assess summary faithfulness is to use off-the-shelf Natural Language Inference (NLI; \citealt{maccartney-manning-2009-extended}) models to determine whether a summary is entailed by the source document.
NLI models determine the semantic relationship between a pair of texts: the \textit{premise} and \textit{hypothesis}. 
If the hypothesis can be inferred from the premise, it is said to be entailed by the premise. 
However, existing NLI models are mainly trained on relatively short texts from existing datasets \citealt{bowman-etal-2015-large,williams-etal-2018-broad}. Examples in these datasets often represent inference cases over a single content unit (e.g., the example at the bottom of  Figure~\ref{fig:intro_ex} where inference is about the \textit{transmission event}). This raises the question of how to apply them to produce entailment judgements for document-summary pairs consisting of multiple sentences aggregating several content units (e.g., the summary sentence MSS in Figure~\ref{fig:intro_ex} aggregates content about the \textit{company launching a legal action}, \textit{a strike event}, and \textit{the consequences of the strike}). Producing an entailment judgement for a summary sentence with several content units is a more complex entailment reasoning task.

\begin{figure*}[t]
    \centering
    \scriptsize
    \begin{tabular}{c|p{10.5cm}ccc}
    \thickhline
&\centering D & $\models$ MSS & $\models$ SS$_1$ & $\models$ SS$_2$  \\
1 &  
\textcolor{cyan}{Lufthansa lost an appeal to a Frankfurt labour court, but is making a further legal challenge that could go late into Tuesday evening.}&0.37&	7.81&	0.06\\

2& \textcolor{cyan}{The pilots' strike}, called over a pay dispute, will affect around 100,000 passengers, Lufthansa said.&0.61	&0.69&	1.74\\

3& The industrial action is part of a long-running pay dispute at Lufthansa. &0.18&	0.74&	0.06\\

4& The pilots' union Vereinigung Cockpit (VC) has organised 14 strikes since April 2014. & 0.07&	0.11&	0.10\\

5& Short and medium-haul flights from Germany will be affected from 00:01 to 23:59 local time (23:01-22:59 GMT). &0.09	&0.14	&0.06\\

6& Flights by Lufthansa's other airlines including Eurowings, Swiss, Austrian Airlines, Air Dolomiti and Brussels Airlines are not affected by the strike, the airline said. &0.11	&0.22	&0.12\\

7& Pay talks between the Vereinigung union and the German airline broke down earlier this month, and Lufthansa said the union had "consistently rejected the offer" of mediation.&0.20	&0.41&	0.06\\

8& The union is calling for a 3.7\% pay rise for 5,400 pilots dating back to 2012. &0.14&	0.42	&0.05\\

9& Lufthansa, which is facing increasing competition from budget rivals, offered a 2.5\% increase over the six years until 2019. &0.12	&0.22	&0.11\\

10& Meanwhile, a separate dispute with cabin crew at Lufthansa's low-cost subsidiary, Eurowings, led it to cancel more than 60 flights on Tuesday.&
0.27&	0.47&	0.32\\

    \hline
        MSS & \multicolumn{4}{p{13cm}}{\textcolor{cyan}{German airline Lufthansa has launched a fresh legal challenge against a strike by its pilots, } \textcolor{red}{which could lead to the cancellation of more than 1,000 flights.}}\\
        
    \hline
        SS & 
\multicolumn{4}{p{13cm}}{\textcolor{cyan}{German airline Lufthansa has launched a fresh legal challenge against a strike by its pilots.}}\\
& \textcolor{red}{The strike could lead to the cancellation of more than 1,000 flights.}\\
      
    \thickhline
    \end{tabular}  
    
    \vspace{0.2cm}
    
    \begin{tabular}{p{6cm}cp{6cm}}
    \thickhline
     At 8:34, the Boston Center controller received a third transmission from American 11 & $\models$ & The Boston Center controller got a third transmission from American 11.\\
     \thickhline
    \end{tabular}    
    
    \caption{Example of input Document (D) and Model-generated Summary Sentence (MSS) from the AggreFact \cite{aggrefact}  benchmark on the XSum \cite{narayan-etal-2018-dont} dataset. The example is considered unfaithful by the annotators. Simplified Summary (SS) is the generated summary after automatic sentence splitting. The \textcolor{cyan}{cyan} coulored text spans in the input document highlight those document content units that support the corresponding \textcolor{cyan}{cyan} spans in the summary. \textcolor{red}{Red} spans in the summary indicate content that is not supported by the input document. The $\models$ MSS and $\models$ SS$_i$ columns show entailment scores assigned by an off-the-shelf NLI model to document sentences acting as premises and either MSS or SS$_i$ sentences as hypotheses. The table in the bottom shows an example of entailment relation from the MNLI dataset \cite{williams-etal-2018-broad}. Entailment scores are computed by the NLI model introduced in Section~\ref{sec:experimental} and normalised for better reading.
    }
    \label{fig:intro_ex}
\end{figure*}

Taking summary sentences as hypotheses, existing approaches try to either identify a document sentence that acts as the premise leading to the highest possible entailment score (sentence-level NLI, \cite{laban-etal-2022-summac,nie-etal-2020-adversarial}) or directly measure entailment by taking the entire document as premise (document-level NLI, \cite{maynez-etal-2020-faithfulness,honovich-etal-2022-true,dziri-etal-2022-evaluating}). However, due to content aggregation happening in summarisation, one document sentence will not contain enough content to entail a summary sentence. In Figure~\ref{fig:intro_ex}, none of the document sentences alone can entail the complex summary sentence MSS aggregating several content units. On the other hand, taking the entire document as premise will perform poorly on long input documents \cite{sentli}).
Recent work achieves promising results by first selecting an entailing context (context-level NLI, \cite{Nie_Chen_Bansal_2019,sentli,wice}).
That is, borrowing insights from information retrieval, these approaches carry out an initial step of document sentence retrieval to build a short context;
and then perform NLI with the retrieved context as a premise. 
Specifically, in the retrieval step, given a summary sentence as hypothesis, document sentences are individually scored by an NLI model and ranked and the top $k$ thereof constitute the premise (e.g., for the MSS in Figure~\ref{fig:intro_ex}, the 2$^{nd}$, 1$^{st}$, and 10$^{th}$ would be selected as premise if $k=3$).

In this work we argue that existing NLI-based approaches do not operate at the right level of granularity \cite{pyramid}; even context-level NLI approaches.
Summary sentences may convey several content units \cite{pyramid} partly overlapping with different document sentences. This renders the retrieval step of document sentences based on NLI scores less accurate (e.g., each document sentence in Figure~\ref{fig:intro_ex} weakly entails the complex summary sentence MSS).  
In addition, summary sentences may aggregate content from different numbers of document sentences which makes it less accurate to have an entailing context with a fixed $k$ number of document sentences (e.g., in Figure~\ref{fig:intro_ex}, SS$_1$ is entailed by two document sentences while SS$_2$ requires only one document sentence to show that its content is not derived from the document).\footnote{Note that {\small{\textit{more than 1,000 flights}}} is not supported by the explicit facts stated in the input document.}
Finally, a fine-grained assessment of summary faithfulness brings interpretability, which hugely facilitates manual inspection of model-generated summaries.

We propose \mbox{\ours}, a faithfulness evaluation approach that \textsc{\textbf{In}}crementally reasons over a document so as to arrive at a \textsc{\textbf{F}}aithf\textsc{\textbf{u}}lnes\textsc{\textbf{s}} \textsc{\textbf{E}}stimation of its summary. 
It aims at retrieving the best possible context to assess the faithfulness of each summary sentence (and in turn the entire summary), i.e., a context with the minimal and most relevant set of document sentences.
Our incremental reasoning approach approximates this via successive expansions of the context adding document sentences and evaluating whether the hypothesis is entailed by it. 
Our approach further decomposes summary sentences for their faithfulness analysis. It does this via sentence simplification. That is, it splits long summary sentences (e.g., MSS sentence in Figure~\ref{fig:intro_ex}) into a set of shorter ones conveying the same content units (e.g., SS$_1$ and SS$_1$ in Figure~\ref{fig:intro_ex}).

Most of previous work focuses on the meta-evaluation of NLI-based approaches on single document news summarisation \cite{laban-etal-2022-summac,aggrefact}. 
Thus, the question of how NLI-based evaluation works on diverse summarisation tasks is left unanswered. 
Hence, to widen the spectrum of NLI-based meta-evaluation \cite{gehrmann-etal-2021-gem}, 
we analyse the performance of NLI-based faithfulness evaluation approaches 
on long document summarisation with diverse domains and genres \cite{cohan-etal-2018-discourse,huang-etal-2021-efficient,zhong-etal-2021-qmsum,adams-etal-2023-desired} and multi-document summarisation \cite{fabbri-etal-2019-multi}.
We collect human annotated model-generated summaries from previous work on these tasks \cite{koh-etal-2022-far,adams-etal-2023-desired,chen-etal-2023-unisumm}. We call this new set the DiverSumm benchmark.

We study existing NLI-based approaches on AggreFact \cite{aggrefact}, a benchmark for the meta-evaluation of single document summarisation, and DiverSumm. \ours~ achieves the best performance in these benchmarks. We find that the choice of an adequate level of granularity for the premise and hypothesis leads to more accurate entailment judgements when using off-the-shelf NLI models. On summaries of extractive nature, retrieving a small relevant set of document sentences suffices. Moreover, our results show that this is crucial for summarisation tasks with long input documents. Summary sentence splitting helps to obtain better performance in all summarisation tasks.

\section{Faithfulness Annotated Data for Different Summarisation Tasks}
\label{sec:faithfulness_sum_tasks}

\begin{table*}[t]
    \small
    \centering
    \setlength{\tabcolsep}{4pt}
    \begin{tabular}{l|cccccp{4.1cm}}
        \hline
         Summarisation Task  &  Doc.Tok & Sum.Sent& Sum.Tok & Cov & Dens & Summarisers \\
        % &&&&Cov  & Dens \\
         \hline
         XSum \cite{aggrefact} & \hspace{0.3em}360.54&1.01& \hspace{0.3em}20.09 & 0.55 & \hspace{0.4em}0.99 & \scriptsize{\textsc{Old-EXformer}, T5, BART, PEGASUS} \\
         CNNDM \cite{aggrefact}& \hspace{0.3em}518.85 & 2.72 & \hspace{0.4em}52.21 & 0.80 & 10.40 & \scriptsize{\textsc{Old-EXformer}, T5, BART, PEGASUS}\\
         ChemSumm \cite{adams-etal-2023-desired}  &  4612.40 & 7.36 &172.79&0.91 & 10.89
         & \scriptsize{LongT5, PRIMERA}\\
         QMSUM \cite{zhong-etal-2021-qmsum}  &  1138.73 & 3.04 & \hspace{0.3em} 65.22 & 0.69 & \hspace{0.4em}5.13 & \scriptsize{GPT-3.5, UniSumm, PEGASUS}\\
         ArXiv \cite{cohan-etal-2018-discourse} & 4406.99 & 6.18 & 149.70 & 0.89 & \hspace{0.4em}9.59 & \scriptsize{PEGASUS, BART}\\
         GovReport \cite{huang-etal-2021-efficient} &2008.16& \hspace{-0.3em}15.07 & 391.22 & 0.86 & 12.76 & \scriptsize{PEGASUS, BART}\\
         MultiNews \cite{fabbri-etal-2019-multi} & \hspace{0.3em}669.20 & 6.81 & 152.20 & 0.82 &14.19 & \scriptsize{GPT-3.5, UniSumm, PEGASUS}\\
         \hline
    \end{tabular}
    \vspace{-0.5em}
    \caption{Statistics on AggreFact (test split) and DiverSumm per summarisation task. Document length in average number of tokens (Doc.Tok), summary length in average number of sentences (Sum.Sent) and tokens (Sum.Tok), and extractive metrics \cite{grusky-etal-2018-newsroom} Density (Dens) and Coverage (Cov). Models generating summaries are LongT5 \cite{guo-etal-2022-longt5}, PRIMERA \cite{xiao-etal-2022-primera}, GPT-3.5 (text-davinci-002) \cite{GPT3,ouyang2022training}, UniSumm \cite{chen-etal-2023-unisumm}, PEGASUS \cite{pegasus}, BART \cite{lewis-etal-2020-bart}, and T5 \cite{2020t5}. As grouped by \citeauthor{aggrefact} (\citeyear{aggrefact}), \textsc{Old-EXformer} denotes older models \cite{see-etal-2017-get,gehrmann-etal-2018-bottom,liu-lapata-2019-text,radford2019language} .}
    \label{tab:aggrefact_diversumm_stats}
\end{table*}

Following previous work, we study faithfulness evaluation on two single document summarisation tasks, namely CNNDM \cite{nallapati-etal-2016-abstractive} and XSum \cite{narayan-etal-2018-dont}. For this, we take the latest introduced faithfulness benchmark, AggreFact \cite{aggrefact}. It consists of a collection of document and model-generated summary pairs where summaries are annotated with faithfulness judgements by human judges. That is, each example in the benchmark is a triple (document, generated-summary, faithful/unfaithful label).
AggreFact includes five annotated sets from the earlier SummaC \cite{laban-etal-2022-summac} benchmark. 
These are XSumFaith \cite{maynez-etal-2020-faithfulness}, FactCC \cite{kryscinski-etal-2020-evaluating}, SummEval \cite{fabbri-etal-2021-summeval}, FRANK \cite{DBLP:conf/naacl/PagnoniBT21}, and Polytope \cite{huang-etal-2020-achieved}. In addition,  AggreFact includes four sets, namely QAGS \cite{wang-etal-2020-asking} (referred as Wang'20 in the benchmark), CLIFF \cite{DBLP:conf/emnlp/Cao021}, GOYAL'21 \cite{goyal-durrett-2021-annotating} and CAO'22 \cite{cao-etal-2022-hallucinated}. AggreFact organises the annotated data into two major sets per summarisation task, CNNDM and XSum, herein we name them CNNDM$_{AG}$ and XSum$_{AG}$. See Appendix~\ref{appendix:datasets} for details on the faithfulness annotation scheme of each dataset and the standarisation criteria applied to derive AggreFact.

\paragraph{DiverSumm a New Benchmark} To study the performance of NLI-based faithfulness evaluation on diverse summarisation tasks, we propose a new benchmark, namely DiverSumm.
It incorporates model generated summaries with human annotations about faithfulness from previous work \cite{koh-etal-2022-far,adams-etal-2023-desired,chen-etal-2023-unisumm}. We follow \cite{laban-etal-2022-summac} to standardise summary annotations into faithful/unfaithful labels. We discuss the summarisation task and characteristics of the annotated sets below.

\begin{enumerate}\itemsep0pt
\item[] \textbf{ChemSumm} \cite{adams-etal-2023-desired} embodies the task of scientific long-form summarisation in the chemistry domain.
Derived from academic journals, each input document contains section headers and associated paragraphs for all sections from the introduction up to the conclusion, and abstracts constitute the reference summaries.

\item[] \textbf{MultiNews} \cite{fabbri-etal-2019-multi} is a large-scale multi-document news summarisation dataset with the number of input documents per example ranging from 2 to 6 and reference summaries written by editors.

\item[] \textbf{QMSUM}  \cite{zhong-etal-2021-qmsum} is a query-based multi-domain meeting summarisation dataset. It consists of meeting transcripts and queries associated with their corresponding abstractive summaries.

\item[] \textbf{ArXiv} \cite{cohan-etal-2018-discourse} is a long scientific paper summarisation dataset collected from ArXiv covering a wide range of topics. The main content up to the conclusion section of a paper is regarded as the document and the corresponding abstract section as the summary.  

\item[] \textbf{GovReport} \cite{huang-etal-2021-efficient} pairs long reports from government research agencies, including the Congressional Research Service and U.S. Government Accountability Office, with expert-written abstractive summaries.

\end{enumerate}

Each summary in QMSUM and MultiNews was labeled using a 5-point Likert scale in terms of fluency, coherence, consistency, and relevance \cite{chen-etal-2023-unisumm}. We use the consistency criterion and label summaries as faithful if the score in consistency is 5, otherwise unfaithful.
In ChemSumm, arXiv, and GovReport, summaries are annotated with a numerical number between 0 (inconsistent) and 1 (consistent) \cite{koh-etal-2022-far,adams-etal-2023-desired}. We take summaries as faithful if the majority of the annotators labeled the summary as~1.

DiverSumm contains 563 test instances with a total of 4686 summary sentences of which 3138 have sentence level annotations.
Table~\ref{tab:aggrefact_diversumm_stats} shows relevant statistics about the benchmarks. Documents and summaries are longer in DiverSumm. Generated summaries for XSum and QMSUM are more abstractive (i.e., smaller coverage and density). 

\paragraph{Error types} 

Some subsets in AggreFact and DiverSumm, namely  FRANK, ArXiv, and GovReport, contain sentence level and detailed error annotations for unfaithful summaries.\footnote{After manual inspection of the human annotations, we filtered out some examples in ArXiv and GovReport with a mismatch between the sentence and summary level annotation.} 
We exploit these annotations to analyse the performance of both the studied approaches and the NLI model on detecting different types of faithfulness errors.
Concretely, unfaithful summaries are annotated with the following error types \cite{DBLP:conf/naacl/PagnoniBT21}.
Relation Error (\emph{PreE}) is when the predicate in a summary sentence is inconsistent with respect to the document. Entity Error (\emph{EntE}) is when the primary arguments of the predicate are incorrect. Circumstance Error (\emph{CircE}) is when the predicate’s circumstantial information (i.e., name or time) is wrong. Co-reference error (\emph{CorefE}) is when there is a pronoun or reference with an incorrect or non-existing antecedent. Discourse Link Error (\emph{LinkE}) is when multiple sentences are incorrectly linked. Out of Article Error (\emph{OutE}) is when the piece of summary contains information not present in the document. Grammatical Error(\emph{GramE}) indicates the existence of unreadable sentences due to grammatical errors.

\subsection{The Value of Adequate Premise and Hypothesis Granularity}
\label{sec:valueof}

\label{sec:rewrite}
\begin{figure}
    \centering
    \includegraphics[width=7.0cm]{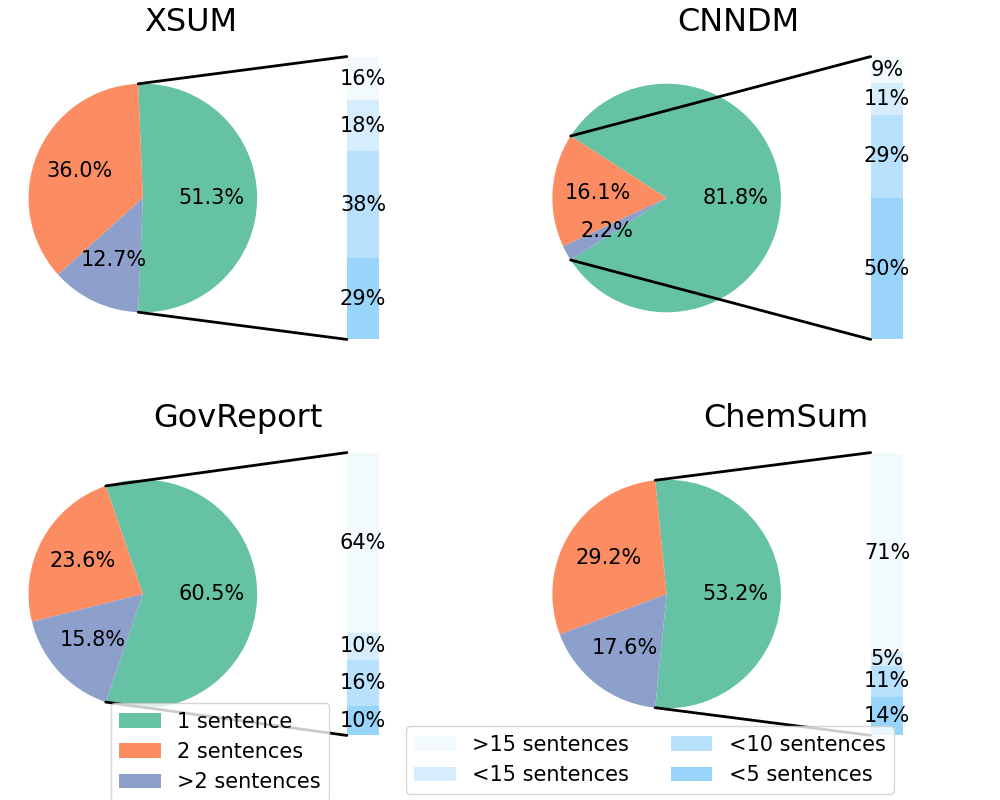}
    \vspace{-0.4em}
    \caption{Statistics for the number of fused document sentences (the pie charts) and their distances (the blue vertical bars) on XSum and CNNDM (AggreFact) and GovReport and ChemSum (DiverSumm).
    }
    \label{fig:sentence_fusion}
\end{figure}

We analyse document-summary pairs in the AggreFact and DiverSumm benchmarks to uncover the rational of why adequate premise and hypothesis granularity brings value into the evaluation of summary faithfulness \cite{Nie_Chen_Bansal_2019,sentli,wice}.

We examine the number of document sentences aggregated into a summary sentence via a greedy selection algorithm that maximizes document-summary token overlap \cite{lebanoff-etal-2019-scoring}. As shown in Figure~\ref{fig:sentence_fusion}, 18-48\% of summary sentences fuse more than one document sentence and at least 50\% of the cases are not within a 5-sentence window. In particular, in GovReport 64\% and ChemSumm 71\% of the times the fused document sentences are in a 15 sentences or more window size. 
This renders sentence- and paragraph-level premises not ideal due to low recall. We show sentence fusion statistics for the other datasets in Figure~\ref{fig:sentence_fusion_appendix}, Appendix~\ref{appendix:datasets}. 

An alternative to improve recall would be via increasing premise size. However, NLI models are typically trained on short premise-hypothesis examples with a premise average length ranging on 16-80 tokens for widely used datasets and a hypotheses length of 9-19  \cite{sentli}.
It is challenging for such models to generalise to document-level premises (average length is 439 in AggreFact and 2566 in DiverSumm). 
Previous work has shown that the performance of faithfulness evaluation consistently drops with longer premises  \cite{sentli}. 
We next describe our approach with premises of variable size (i.e, variable number of document sentences) and shorter hypotheses (i.e., simplified summary sentences).

\section{\ours}
\label{sec:infuse}

We denote a document as $D=\{d_m\}_{m=1}^M$ and a summary as $S=\{s_n\}_{n=1}^N$ where $d_m$ and $s_n$ are sentences. 
For a given summary sentence $s_n$ as the \textit{hypothesis}, 
we aim to retrieve a related context $R^{(n)}$, $R^{(n)} \subseteq D$, to act as the \textit{premise} and estimate whether $s_n$ can be entailed by $R^{(n)}$ (and, therefore, $D$) according to an NLI model $\theta$. We assume that $\theta$ predicts one of the $\{\text{entailment}, \text{neutral}, \text{contradict}\}$ labels for a given premise-hypothesis pair.
Summary sentence faithfulness estimates, given by $\theta(\text{entailment}|\cdot)$,
%of faithfulness 
are then aggregated into summary faithfulness scores with mean pooling.

\paragraph{Incremental Reasoning}

Given an NLI model $\theta$, we construct a matrix $E$ of entailment scores via sentence level inference between document sentences $d_m$ and each summary sentence $s_n$. 
We derive from $E$ entailment ranked lists of document sentences $\hat{D}^{(n)}$ 
associated to each summary sentence $s_n$. 
%We then perform incremental context retrieval. 
We then incrementally select sentences from $\hat{D}^{(n)}$ in a top-down fashion to retrieve a context $R^{(n)}$ for $s_n$. Starting from an empty context $R^{(n)}_0$, at each step $i$, we remove the top sentence from $\hat{D}^{(n)}$ and incorporate it into the current context to obtain a new context $R^{(n)}_i$. We then stop adding sentences to the context when the local minimum of the neutral class probability, $u_{i,n} = \theta(\text{neutral}|R^{(n)}_{i},s_n)$, is reached, i.e., $u_{i,n} \ge u_{i-1,n}$.
Intuitively, decreasing neutral scores signal shifts in the perceived entailment relationship from context $R^{(n)}_{i-1}$ to $R^{(n)}_{i}$ (i.e., candidate premises) and $s_n$ (the hypothesis) leaning towards either entailment or contradiction. We stop when there is an increase in the neutral score. 
At this stopping point, the entailment score between the premise given by context $R^{(n)}_{i}$ and summary sentence $s_n$ as hypothesis is taken as the final faithfulness estimation for $s_n$.

\begin{small}
\begin{algorithm}
\caption{Summary sentence entailment estimation in \ours.}
\begin{algorithmic}[1]
\Require NLI model \(\theta\), pair $(D, s_n)$.
\Ensure $R^{(n)}_{i-1}, e_{i-1,n}$ premise and entailment score for $s_n$.

\For{\( d_{m} \) in $D$ }
\State $e_{m,n}, u_{m,n}, c_{m,n} = \theta(d_m,s_n) $
\State $e_{n,m}, u_{n,m}, c_{n,m} =  \theta(s_n,d_m)$ 
\State {\small \# entailment \(e \), neutral \(n \), contradiction \(c \)}
\State $\hat{E}{_{d_m,s_n}} = e_{m,n} + e_{n,m}$ 
\EndFor

    \State $\hat{D}^{(n)} = rank(\hat{E}{_d{_{1:M}},s_n})$
    \State $R^{(n)}_0 = \emptyset, n_{0,n} = 0 $
    \For{$\hat{d}_{i}$ in $\hat{D}^{(n)}$}
    \State add $\hat{d}_{i}$ to $R^{(n)}_i$    
    \State $e_{i,n}, u_{i,n}, c_{i,n} = \theta(R^{(n)}_i, s_n) $
    \If{  \(u_{i,n} \ge u_{i-1,n} \)} 
     \State stop and return $R^{(n)}_{i-1}, e_{i-1,n}$
    \EndIf
    \EndFor
\end{algorithmic}
\label{alg:infuse}
\end{algorithm}

\end{small}

\paragraph{Reversed Reasoning} 

In some cases, the content expressed in a document sentence $d_m$ will only entail part of a summary sentence $s_n$ 
(see example in Table~\ref{tab:effects_of_reverse} -bottom- of the Appendix). Thus, such $d_m$ will have a low sentence level entailment score in $E$ despite $d_m$ really providing evidence for a part of $s_n$.
Because summaries will contain extracted document fragments or paraphrases thereof,
one way to improve entailment scores for such document sentences $d_m$ is to reverse the direction in which sentence level NLI is applied. That is, we take the summary sentence $s_n$ as premise and the document sentence $d_m$ as hypothesis. We add reversed entailment scores to those on $E$ and obtain a new re-weighted matrix $\hat{E}$ which is adopted to perform incremental context retrieval. Algorithm~\ref{alg:infuse} summarises \ours~ steps to estimate the entailment score of a given summary sentence with respect to its corresponding input document.

\paragraph{Sub-sentence Reasoning}
\label{sec:rewrite}
Different document sentences $d_m$ will entail different parts of a summary sentence $s_n$  (see document sentence fusion in Figure~\ref{fig:sentence_fusion}). In addition, those document sentences  $d_m$ may contain irrelevant content for $s_n$. Thus, sentence level scores in $E$ as well as final context level entailment scores for $s_n$ will be noisy (i.e., more chances of having neutral class high scores).
Shorter summary sentences with finer-grained content units will yield more accurate contexts and entailment estimations. Figure~\ref{fig:intro_ex} illustrates the difference in entailment scores in $E$ when computed on the original summary sentence (MSS) and when computed on its sub-sentences (SS$_1$ and SS$_2$).
In this work, we propose to simplify each summary sentence by splitting it into multiple sub-sentences.

\begin{table*}[t]
\centering
\footnotesize
\setlength{\tabcolsep}{2pt}
\def\arraystretch{1.2}
\begin{tabular}{lcccccccc}
\thickhline
 & \multicolumn{1}{c}{XSM$_\textsc{ag}$} & \multicolumn{1}{c}{CND$_\textsc{ag}$} & \multicolumn{1}{c}{CSM} & \multicolumn{1}{c}{QMS} & \multicolumn{1}{c}{AXV} & \multicolumn{1}{c}{GOV} &   \multicolumn{1}{c}{MNW} & \multicolumn{1}{c}{AVG}  \\
\hline
\fulldoc & 72.77 & 64.40 & 50.15 &  37.12 & 62.78 & 79.19 & 44.76 & 58.74 \\

\summac$_\textsc{conv}$  & 67.76&	72.14&	53.14&	51.13&	61.22&	65.34&	53.05&60.54  \\

\summac$_\textsc{zs}$ & 70.29&	74.54 & 54.41&	48.21&	69.44&	79.37&	50.17&	63.78 \\

\sentli  & \textbf{73.61} & 75.83 & 50.13 & 47.56 & 64.49 & 79.68 & 46.61 & 62.56 \\

\hline 

\ours & 73.42 & \textbf{76.21} & 54.11 & \underline{52.16} & \underline{71.38} & \textbf{80.45} & \textbf{53.16} & \textbf{\underline{65.84}} \\

\oursub  & 73.21 &	73.34 & \textbf{\underline{59.26}} & \textbf{\underline{53.20}} & \textbf{\underline{73.89}} & 80.05 & 49.37 & \textbf{\underline{66.05}}  \\ % this is variant -split (min+avg) 

\thickhline
\end{tabular}
\vspace{-0.5em}
\caption{\label{tab:res_aggrefact_and_diversumm}
Results for all summarisation tasks in AggreFact and DiverSumm. 
For AggreFact, we report the average results for XSum (XSM; 5 datasets) and CNN/DM (CND; 7 datasets), respectively; dataset-level performance can be found in Appendix \ref{appendix:more_results}. 
CSM, MNW, QMS, AXV, and GOR refer to ChemSum, MultiNews, QMSUM, ArXiv, and GovReport respectively. 
We highlight \textbf{highest} scores and scores \underline{significantly different} from \fulldoc,  \textsc{SummaC} variants, and \sentli~ approaches (at $p<.05$).
\vspace{-1.5em}
}

\end{table*}

\section{Experimental Setup}
\label{sec:experimental}

We study NLI-based faithfulness evaluation approaches on AggreFact \cite{aggrefact} and DiverSumm (Section~\ref{sec:faithfulness_sum_tasks}).
We adopt an {ALBERT-xlarge} \cite{DBLP:conf/iclr/LanCGGSS20} model optimized on MNLI \cite{williams-etal-2018-broad} and VitaminC \cite{schuster-etal-2021-get} as our NLI model $\theta$. 
MNLI covers ten distinct genres and styles of written and spoken English data. It aims to support a broader understanding and analysis of NLI across different genres and domains. 
VitaminC is synthetically created from Wikipedia sentences. Claim sentences are associated with contrastive evidence, i.e., one sentence that supports the claim and another one that does not.
On MNLI (VitaminC) premises are 13.23 (43.03) tokens long in average and hypotheses 13.23 (27.57).

We fine-tune a T5-large \cite{2020t5} model for sentence splitting. We use this model to simplify sentences in model generated summaries. We manually inspect several samples of split sentences and find that the performance is reasonable. Details about our sentence splitting model, examples, and statistics about the percentage of sentence splits are presented in Appendix \ref{appendix:training}. 

We compare \ours~ with a widely adopted approach
%for summarisation evaluation 
which considers the entire document as a premise, we refer to it as \fulldoc. In practice, it divides the input document into chunks that fit the input size of the NLI model, computes chunks scores and takes the average thereof. 
We also compare with \textsc{Summac}$_\textsc{ZS}$ \cite{laban-etal-2022-summac}, a sentence-level method which assumes each summary sentence is supported by one document sentence, 
and takes the one with the top entailment score as context. 
\mbox{\textsc{Summac}$_\textsc{Conv}$} introduces a convolutional layer trained on a subset of FactCC \cite{kryscinski-etal-2020-evaluating} to aggregate the score given by an NLI model to each $\{\text{entailment}, \text{neutral}, \text{contradict}\}$ label into a final score. For a fair comparison with the other models, we remove specific constraints used in the original implementation of \textsc{Summac} variants (see Appendix \ref{appendix:training}).
\sentli~\cite{sentli} retrieves a context with a fixed number $k$ of document sentences. Its context includes document sentences with top entailment and top contradiction scores. Following \cite{sentli}, we set the value of $k=5$. We show performance with other values of $k$ in Figure \ref{fig:premise_size} in Appendix~\ref{app:performance_on_k}.
\oursub~ is our variant~with sub-sentence reasoning (i.e., summary sentence simplification). For this variant, to better mimic the process of label standardisation as described in Section \ref{sec:faithfulness_sum_tasks}, we use the \emph{min($\cdot$)} operator to aggregate the entailment scores from sub-sentences into a sentence score.

\section{Results}
\label{sec:results}

\subsection{Faithfulness Evaluation}
\label{sec:overall_auc}

Following \citet{laban-etal-2022-summac}, we adopt ROC-AUC \citep{bradley1997use} which depicts classification performance with varied thresholds as our evaluation metric. 
Results on AggreFact and DiverSumm are shown in Table \ref{tab:res_aggrefact_and_diversumm}.\footnote{ 
To determine the statistical significance of performance differences, we randomly re-sample 70\% of the test instances 100 times and evaluate the models on these sets. We use the pairwise t-test to assess whether there is a significant difference between models.}
\ours~ and \oursub~ exhibit superior performance than previous approaches overall summarisation tasks. 
\fulldoc~ exhibits the lowest performance, this confirms results from previous meta-evaluations \cite{laban-etal-2022-summac,sentli} and extends the observations to the summarisation tasks in our DiverSumm benchmark.
\summac$_\textsc{conv}$, trained on specific evaluation data, does not generalise well across the different summarisation tasks. Thus, our main comparison variant is \summac$_\textsc{zs}$.

As for the role of sub-sentence reasoning, 
we observe that \oursub~works better on ChemSumm, QMSUM, and ArXiv where summary sentences are complex and informative (see sentence fusion in Figure~\ref{fig:sentence_fusion}) and more abstract (Table~\ref{tab:aggrefact_diversumm_stats}). 
This further validates the positive findings from claim verification tasks \cite{wice} for text summarisation.
On the other hand, sub-sentence reasoning is less effective on CNNDM$_{AG}$, GovReport and MultiNews which consist of more extractive summaries (Table~\ref{tab:aggrefact_diversumm_stats}). In  CNNDM$_{AG}$ segmenting short sentences may only introduce noise.
This partially supports  \citet{DBLP:journals/corr/abs-2211-16853} who draw a negative conclusion on the effectiveness of sub-sentence evaluation based on CNNDM.
We note that the nature of the data underlying evaluation benchmarks should be further emphasized to delimit the scope of conclusions drawn. For GovReport and MultiNews, with the most extractive summaries (Table~\ref{tab:aggrefact_diversumm_stats}), we found that after splitting the relation between summary sub-sentences and document sentences becomes mostly one-to-one and thus approaches taking one document sentence become more effective (see results for existing approaches with sub-sentence hypotheses in Appendix~\ref{app:sub_approaches}).

On XSum$_{AG}$, retrieval approaches, \ours~ and \sentli, work very closely to the document-level approach \fulldoc. The success of the document-level approach lies on the fact that summaries in XSum$_{AG}$ are highly abstractive (Table~\ref{tab:aggrefact_diversumm_stats}) and require reasoning over multiple document sentences; and input documents are short. Indeed, for highly abstractive summarisation tasks such as XSum$_{AG}$  or QMSUM it would make sense to build a structured premise with document sub-sentence content, connecting discourse information, explicit world knowledge, and intermediate inferences made explicit \cite{dalvi-etal-2021-explaining}.

Results in Table~\ref{tab:res_aggrefact_and_diversumm} show that the variable premise size of \ours~ leads to better performance across the board. In Appendix~\ref{app:performance_on_k}, we show performance curves for \ours~ versus \ours$-k$, a version with different fixed premise sizes, to further illustrate this.
We report statistics about the number of document sentences retrieved by \ours~ and \oursub~ in Table \ref{tab:kvalue}. These show the inherent variability
in document sentence fusion happening within summary sentences (see approximation of this in Figure~\ref{fig:sentence_fusion}).

The reversed entailment direction in the retrieval step acts as a re-weighting scheme that takes advantage of paraphrased content and favors shorter document sentences (i.e., fewer content units than those appearing in summary sentences). 
We provide performance curves on the effect of reversed reasoning comparing \ours~ with a version \ours$-reversed$ in Appendix~\ref{app:performance_on_k}; and case studies in Appendix \ref{appendix:analysis} .

On DiverSumm, ROC-AUC scores are considerably lower than those obtained on AggreFact across the board. We attribute this to the summarisation tasks been more complex and recent models that generated the summaries more powerful. The lowest scores are on ChemSumm and QMSUM, we attribute this to the shift in vocabulary and genre in these tasks. The following lowest scoring task is MultiNews, we attribute this to the redundancy found in multi-document input. The fixed context of \sentli~ will only include redundant sentences.

\subsection{Performance on Different Error Types}
\label{sec:performance_error_types}

\begin{figure*}[t]
    \centering
    \begin{subfigure}{.30\textwidth}
    \centering
    \includegraphics[width=1\linewidth]{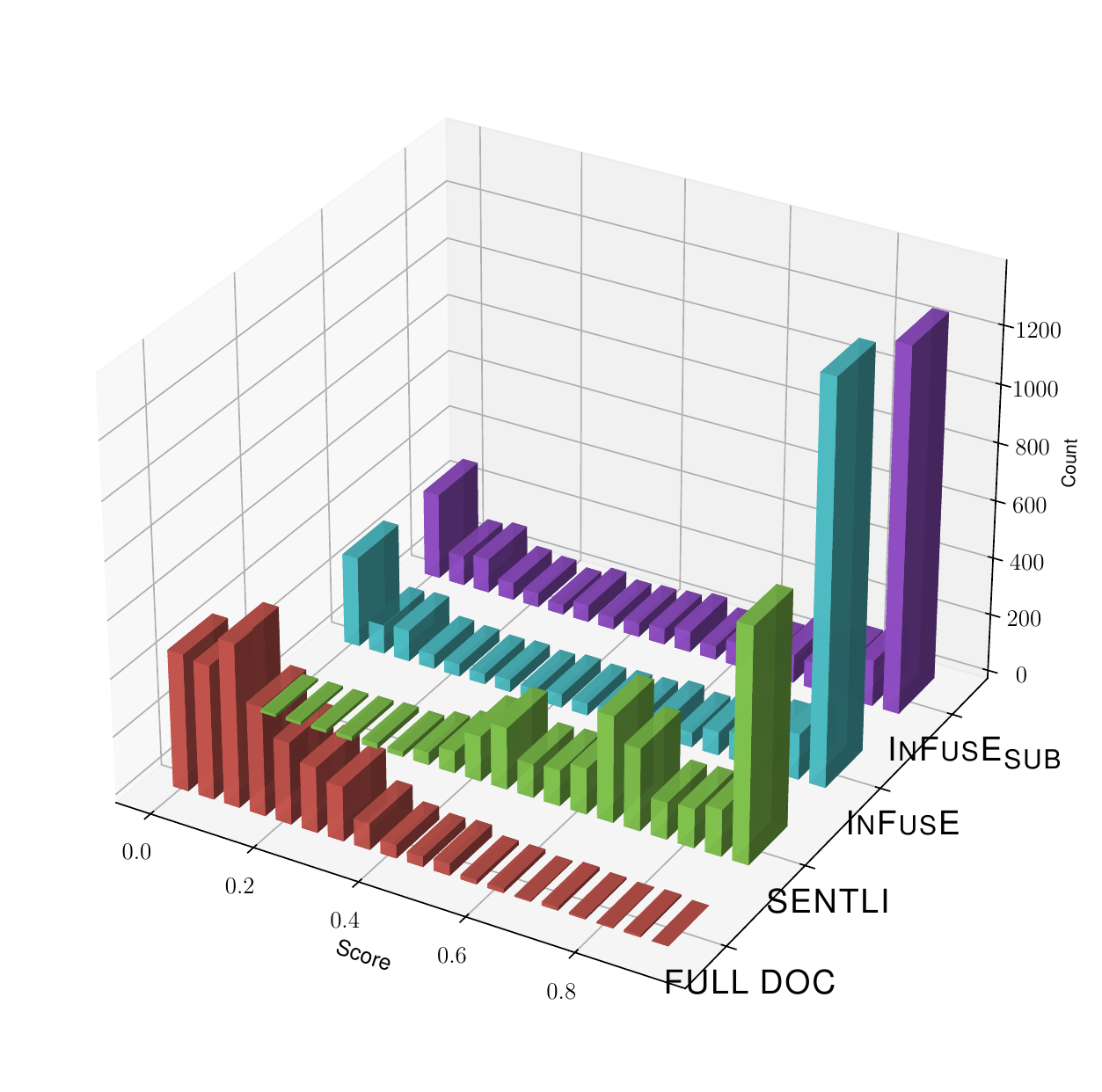}
    \vspace{-0.8cm}
    \caption{Faithful}
    \label{fig:sfig1}
    \end{subfigure}
    \begin{subfigure}{.30\textwidth}
    \centering
    \includegraphics[width=1\linewidth]{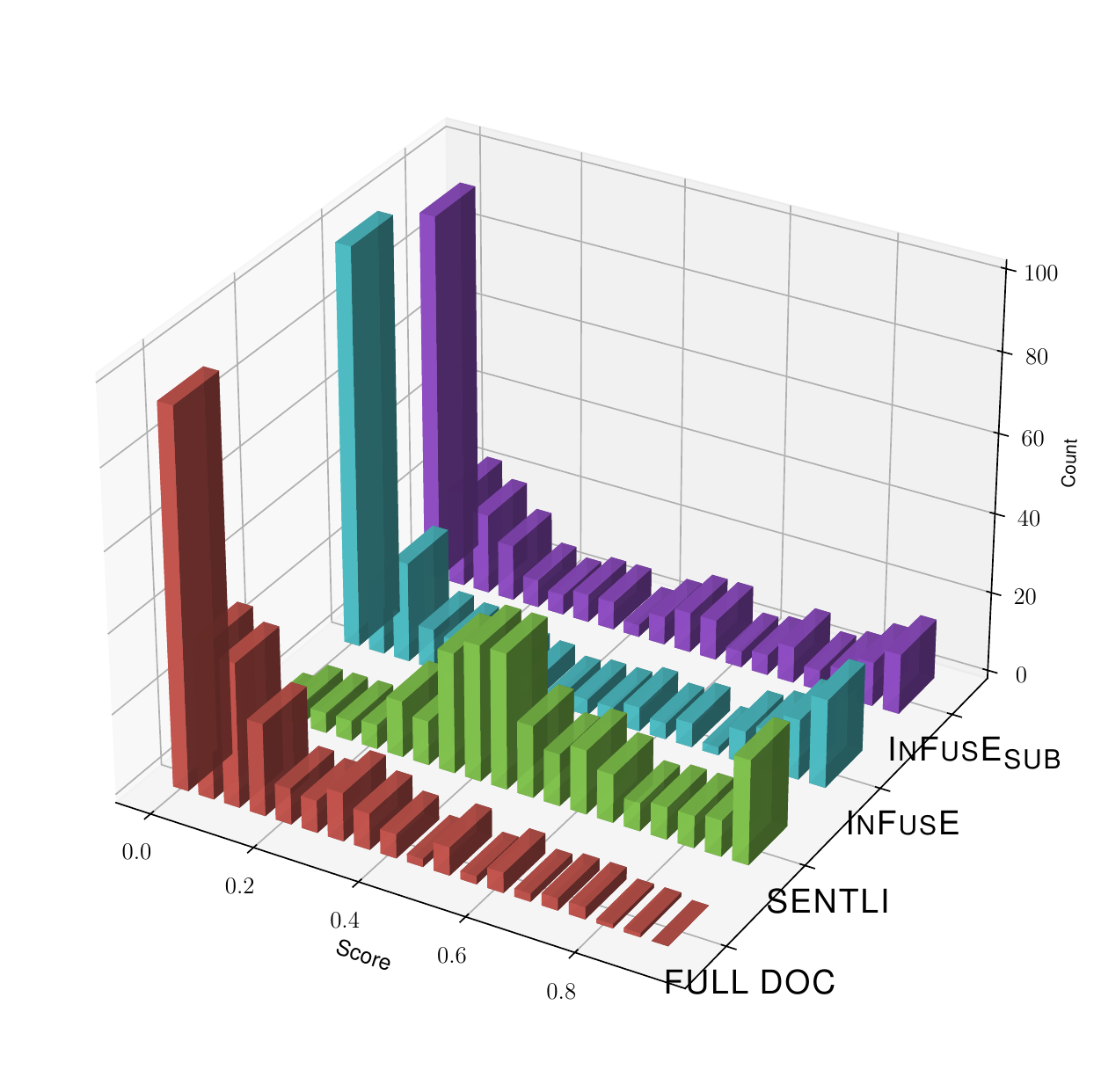}
    \vspace{-0.8cm}
    \caption{EntE}
    \label{fig:sfig2}
    \end{subfigure}
    \begin{subfigure}{.30\textwidth}
    \centering
    \includegraphics[width=1\linewidth]{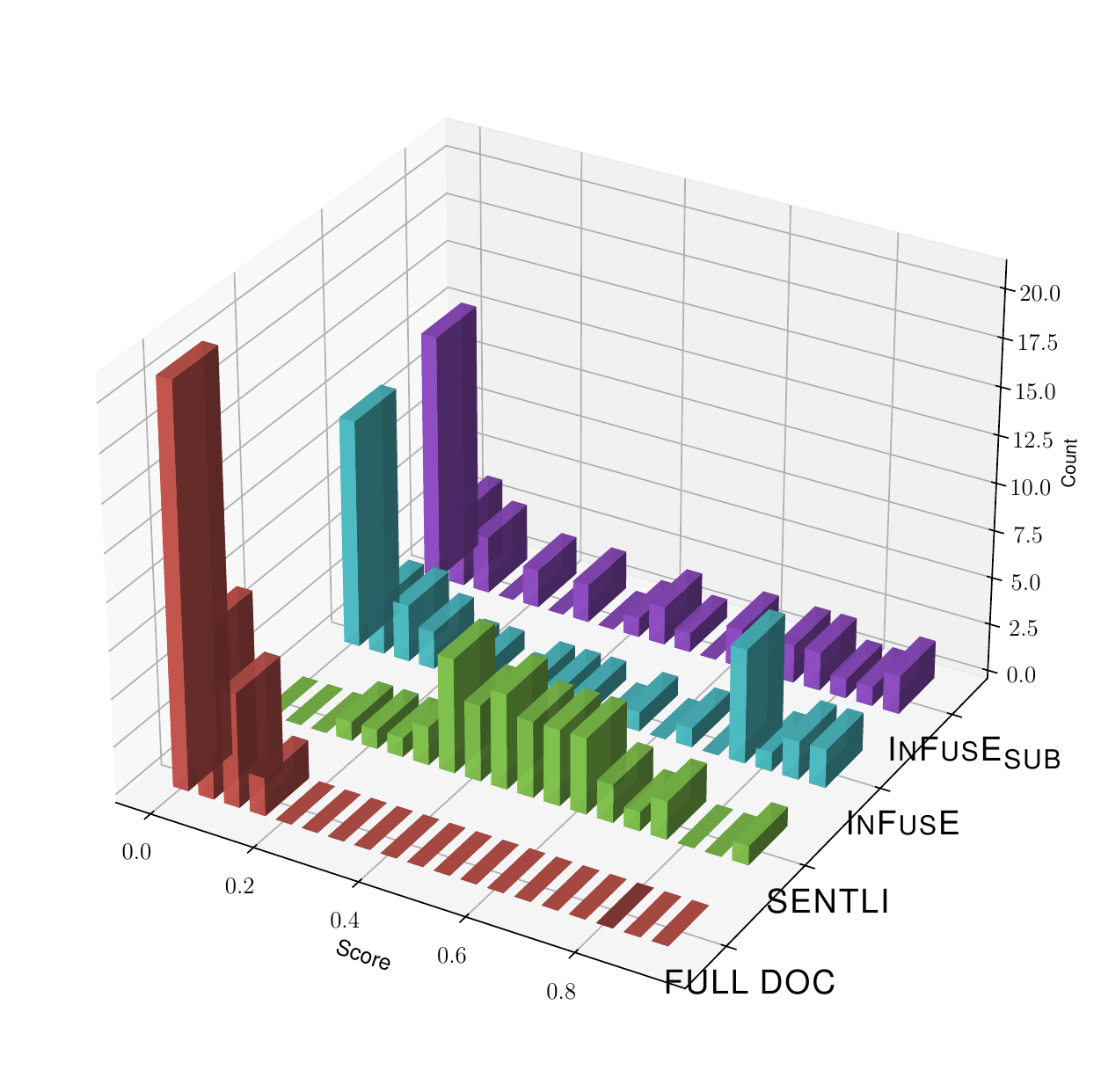}
    \vspace{-0.8cm}
    \caption{LinkE}
    \label{fig:sfig2}
    \end{subfigure}    
    %%%
    \\
    \begin{subfigure}{.30\textwidth}
    \centering
    \includegraphics[width=1\linewidth]{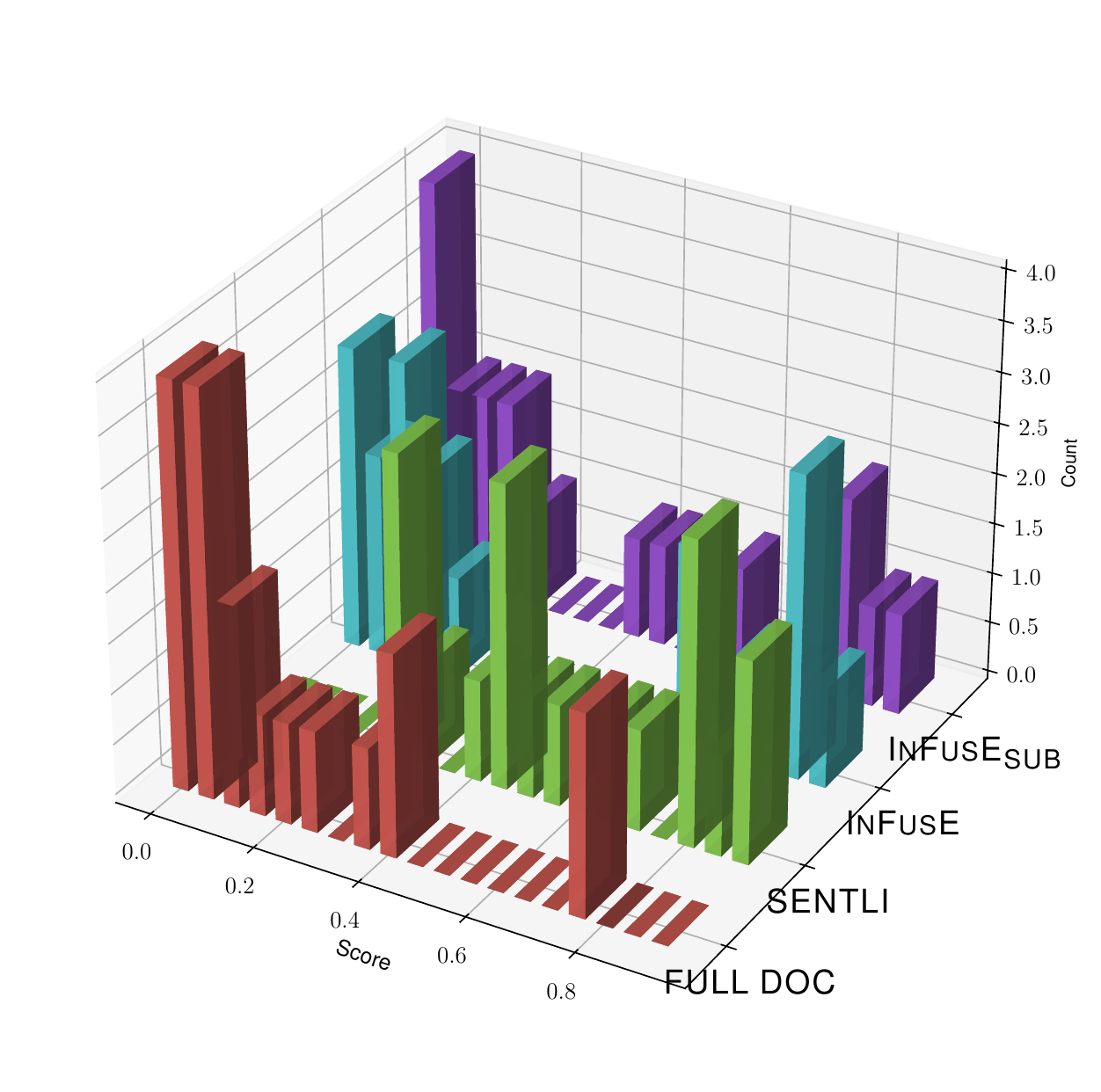}
    \vspace{-0.8cm}
    \caption{CorefE}
    \label{fig:sfig1}
    \end{subfigure}
    \begin{subfigure}{.30\textwidth}
    \centering
    \includegraphics[width=1\linewidth]{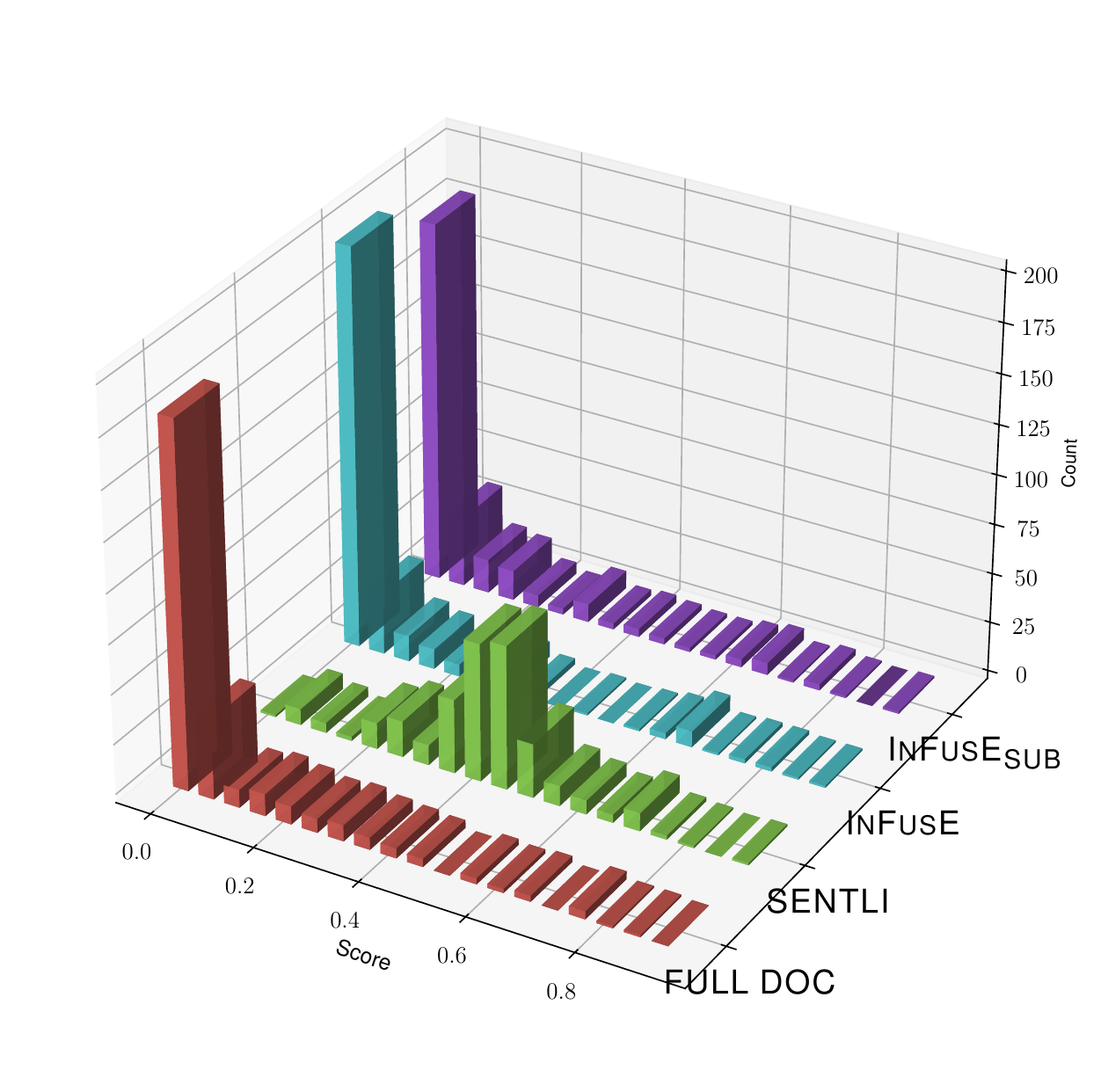}
    \vspace{-0.8cm}
    \caption{OutE}
    \label{fig:sfig3}
    \end{subfigure}
    \begin{subfigure}{.30\textwidth}
    \centering
    \includegraphics[width=1\linewidth]{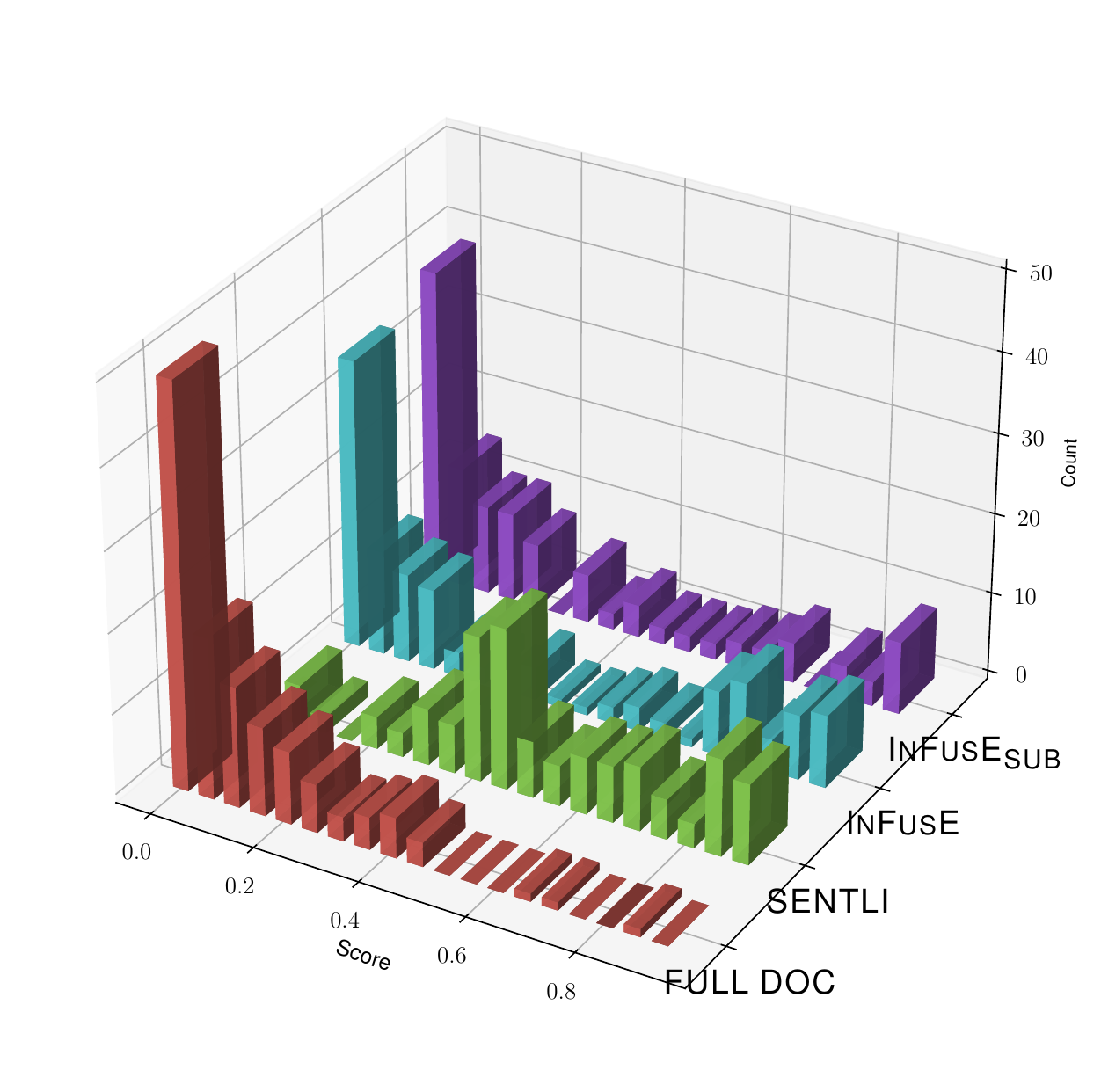}
    \vspace{-0.8cm}
    \caption{GramE}
    \label{fig:sfig2}
    \end{subfigure}
    \caption{Distribution of entailment scores on faithful summary sentences and unfaithful ones encompassing different error types for ArXiv, GovReport and FRANK sets. The x-axe corresponds to the NLI-based approach. That is, \fulldoc~ in red, \sentli~ in green, \ours~ in cyan, and \oursub~ in purple. The y-axe corresponds to the entailment scores (i.e., values ranging in [0,1]), and the z-axe corresponds to the count of instances.
    }
    \label{fig:error_types_ent_dist}
\end{figure*}

We look into the performance of the studied NLI-based approaches with respect to unfaithfulness errors discussed in Section~\ref{sec:faithfulness_sum_tasks}. 
We focus on ArXiv, GovReport, and FRANK which contain fine-grained error annotations at sentence level. We consider each summary sentence in these subsets to be labelled with the error types that the majority of annotators agreed upon.
We analyse the distribution of entailment scores for \fulldoc, \sentli, \ours, and \oursub~ on summary sentences (i.e., without aggregation into a final summary score). We show these in Figure~\ref{fig:error_types_ent_dist}.\footnote{Note that for none of the approaches, we have tuned a faithful/unfaithful decision threshold; however, we compare faithful/unfaithful distributions and analyse performance at extreme scores 1/0 in the [0,1] interval.} 

Before looking into specific error types we analyse the range of scores the approaches assign to faithful sentences.
Figure~\ref{fig:error_types_ent_dist}.a shows that  \fulldoc~ tends to predict rather low entailment scores, close to zero, for most of faithful cases. This explains the lower ROC-AUC in Table~\ref{tab:res_aggrefact_and_diversumm}. Using the entire document leverages noise when computing entailment if the input documents are long. After all, NLI models are trained to use the entire premise to yield a judgement and not to distinguish those relevant from irrelevant premise parts. Context-level approaches produce higher scores. \ours~ and \oursub~ produce more extreme scores.

On EntE error types, Figure~\ref{fig:error_types_ent_dist}.f shows that \ours~ assigns close to zero scores to more EntE cases. It works slightly better than \oursub~ and we attribute this to the fact that \oursub~ may introduce some noise when splitting sentences. In contrast, \sentli~ assigns entailment scores in the range of [0.4,0.6] where also many faithful cases fall on. Figure~\ref{fig:error_types_ent_dist_add}.a/.b in Appendix~\ref{app:analysis_error_type} shows a similar trend for PredE and CircE error types.

As for discourse level errors, on LinkE error types, Figure~\ref{fig:error_types_ent_dist}.c, \ours~ works better. After manual inspection, we attribute this to the fact that none of the incorrectly fused document sentences contributes to a high entailment score. Thus, they will not be retrieved as an entailing premise.
On CorefE errors, Figure~\ref{fig:error_types_ent_dist}.d, we can see that all approaches have poor performance assigning relatively high entailment scores. 
Note that the set of these error types is rather small.
 
Finally, on the OutE error type, Figure~\ref{fig:error_types_ent_dist}.e, \ours~ is better over \oursub~ and \sentli. We attribute this to the fact that in cases of this error type there is no document information that can support nor contradict the summary sentence; thus, \ours~ will take the minimum number of document sentences (potentially only one) failing to entail the summary with OutE.
Grammar errors, GramE in Figure~\ref{fig:error_types_ent_dist}.f, seem difficult to detect, which makes sense for NLI-based approaches.

We observed that in some error types \ours~ (and \oursub) assigns extremely high ($\sim 1$) scores to some cases. We manually examine a sample thereof and find that in most cases summary sentences have a high lexical overlap with document sentences and vary either on few tokens or word order. Thus, the NLI model is biased to rely on extractive cues \cite{mckenna2023sources,verma-etal-2023-evaluating}. 
Table~\ref{tab:high_low_error_types} in Appendix~\ref{app:analysis_error_type} shows examples of error types correctly ($\sim 0$) and incorrectly ($\sim 1$) evaluated by \ours~.

\section{Related work}

Some NLI-based approaches directly train document level NLI models \cite{yin-etal-2021-docnli,utama-etal-2022-falsesum}. Others leverage off-the-shelf NLI models \cite{Nie_Chen_Bansal_2019,nie-etal-2020-adversarial,laban-etal-2022-summac,sentli,wice,steen-etal-2023-little}. The former requires the construction of synthetic training data. In this paper, we study the latter type of approaches. 
These do not require additional data nor training resources.

\cite{nie-etal-2020-adversarial,laban-etal-2022-summac} select a single sentence as premise while \cite{Nie_Chen_Bansal_2019,sentli,wice} select a fixed number of document sentences, the same for all summary sentences. Our approach selects a variable number of document sentences as premise for each summary sentence.
Recently, \cite{10.1162/tacl_a_00576} conduct an empirical analysis of how to use the three directions in which entailment can be computed (entailment direction implication, reverse implication, and bi-implication). However, \cite{10.1162/tacl_a_00576} directly use the scores from these directions in a single pass using the entire document as premise. In contrast, we apply reversed reasoning only to re-weight document sentences in the context retrieval step.
\cite{steen-etal-2023-little} propose a document-level approach with data augmentation to adapt NLI models to task specific scenarios such as dialogue. Furthermore, they ensemble a number of calls to the NLI model via Monte-Carlo dropout to cope with domain shift. These ideas are orthogonal to our work and would make sense to use them in combination.

The value of fine-grained assessment of summary content has been highlighted in earlier work on summarisation evaluation \cite{marcu-duc2001,voorhees,van-halteren-teufel-2003-examining,teufel-van-halteren-2004-evaluating,pyramid,gao-etal-2019-automated,shapira-etal-2019-crowdsourcing}. This research highlights that summary sentences aggregate several content units and judgements should be initially provided for these before yielding a conclusion at summary level. However its focus is on the evaluation of content relevance.
Recent work in the context of summary faithfulness evaluation assesses faithfulness of summary predicates and arguments \cite{goyal-durrett-2021-annotating}.
Conciliating with our results, they also show that fine-grained evaluation is beneficial. However, their approach is not based on NLI; and requires syntactic analysis of summary sentences and task specific human annotated data 
to train a classifier. Our approach is more generic and builds on existing resources.  Contemporary with our work, \cite{min2023factscore} propose the evaluation of Large Language Models (LLMs) generated biographies via their decomposition into smaller content units (i.e., atomic facts). Their approach is applied to factual descriptive generation. In contrast, we evaluate hallucination detection in a variety of summarisation tasks.
For long dialogue summarisation, \cite{LattimerC0Y23} propose to decompose the input into chunks, \ours~ could be combined with a coarse chunk selection step.

Finer-grained evaluation has also shown positive results in the related task of claim verification \cite{chen-etal-2022-generating,wice}. However, in the same way as current factuality evaluation on LLM generated text \cite{min2023factscore,manakul2023selfcheckgpt}, they address more open-ended generation tasks where no ground truth input is assumed; their information source is either retrieved or parametric. We focus on NLI-based faithfulness evaluation 
from given input documents.

\section{Conclusions}

We study existing NLI-based faithfulness evaluation approaches and propose a new one, \ours, that works at finer-grained granularity levels for computing document-summary entailment judgements.
Our study shows that lower granularity via premises with variable size and summary sentence splitting is key to achieve more accurate entailment judgements when using off-the-shelf NLI models.
We also introduce a new benchmark for long form input and diverse summarisation tasks.
Experimental results show that \ours~ achieves superior performance on evaluating faithfulness for diverse summarisation tasks.

\section*{Limitations}

\ours's stopping criteria can fall into a local minimum. In Table~\ref{tab:kvalue} (see Appendix \ref{app:performance_on_k}), we show the average number of document sentences retrieved by {\ours}. It is evident that \ours~ incremental context retrieval 
extracts more document sentences on XSum$_{AG}$ than on CNNDM$_{AG}$.
This aligns with our analysis in Section~\ref{sec:valueof} and the fact that summaries in XSum$_{AG}$ are more abstractive than those in CNNDM$_{AG}$. However, it might still not be enough, especially in XSum$_{AG}$, where some summary sentences indeed require more document sentences to form an entailing context. As a result, {\ours} performance is comparable to {\sentli} which manually sets a fixed number of document sentences to be retrieved.   
This limitation can be overcome by introducing a hyper-parameter to the stopping criterion (Section~\ref{sec:infuse}); for example, to stop expanding the context when the neutral probability increases only by a large margin. 
The stopping criterion we adopt is simple but enough to show that it is possible to improve faithfulness evaluation performance when using off-the-shelf NLI models by allowing a variable premise size.

Another limitation of \ours~ is that it requires additional calls to the NLI model. In Table~\ref{tab:tradeoff}, we show for all the compared approaches the computation cost of evaluating one summary sentence versus the achieved average performance. The reversed reasoning re-weighting in \ours~ doubles the computation cost when compared with \sentli~ and \summac. 
However, in practice, it would be possible to decrease the number of calls by using some heuristics to flag when it is necessary (or not). For instance, when the entailment score is above some threshold the reversed direction is not analysed; or the decision could be based on whether the summary sentence fuses more than one document sentence which can be computed based on a cheap metric such as ROUGE. 
The automatic stopping criteria requires a number of additional calls given by the expected number of retrieved document sentences $k_{avg}$ taken as premise. 
The complexity of inference for a context-level approach with a fixed number of retrieved sentences $k$, i.e., \ours-$k$ or \sentli, assuming a standard transformer, is 
$O(k^2)$ whereas for \ours~ it is in $O(k_{avg}^3)$.
If $k_{avg}$ is small enough and there is variability in the number of retrieved document sentences, which is the case in the analysed summarisation tasks (see  Table~\ref{tab:kvalue} in Appendix~\ref{app:performance_on_k}), \ours~ can be competitive in terms of running times. 
Summary sentence splitting also adds an extra overhead; however, it will decrease summary sentence fusion of document sentences, i.e., fewer cases will need reversed reasoning and $k_{avg}$ will be smaller.
In terms of performance, the contribution of reversed reasoning and dynamic stopping can be seen in Figure~\ref{fig:premise_size} in Appendix~\ref{app:performance_on_k}. Although grid search for $k$ will give the best possible $k$, this $k$ value will be the same for all summary sentences (within a summary and within a dataset). In contrast, dynamic stopping lets each summary sentence be analysed with a different $k$ value. Figure~\ref{fig:premise_size} shows that \ours~ with dynamic stopping is better than \ours-$k$ for different values of $k$.

% Ethics section is optional
% \section*{Ethics Statement} 

\section*{Acknowledgements}

We thank our reviewers for their constructive feedback. We are grateful to Griffin Adams, Huan Koh, Yulong Chen, and Yang Liu for making available the annotated datasets that we include in DiverSumm. We also thank Mirella Lapata for useful feedback. 
We gratefully acknowledge the support of the UK Engineering and Physical Sciences Research Council (grant EP/L016427/1).

% Entries for the entire Anthology, followed by custom entries
\bibliography{eacl2024-infuse}
\bibliographystyle{acl_natbib}

\appendix

\section{Additional Dataset Details}
\label{appendix:datasets}

\begin{figure}[t]
    \centering
    \includegraphics[width=7.0cm]{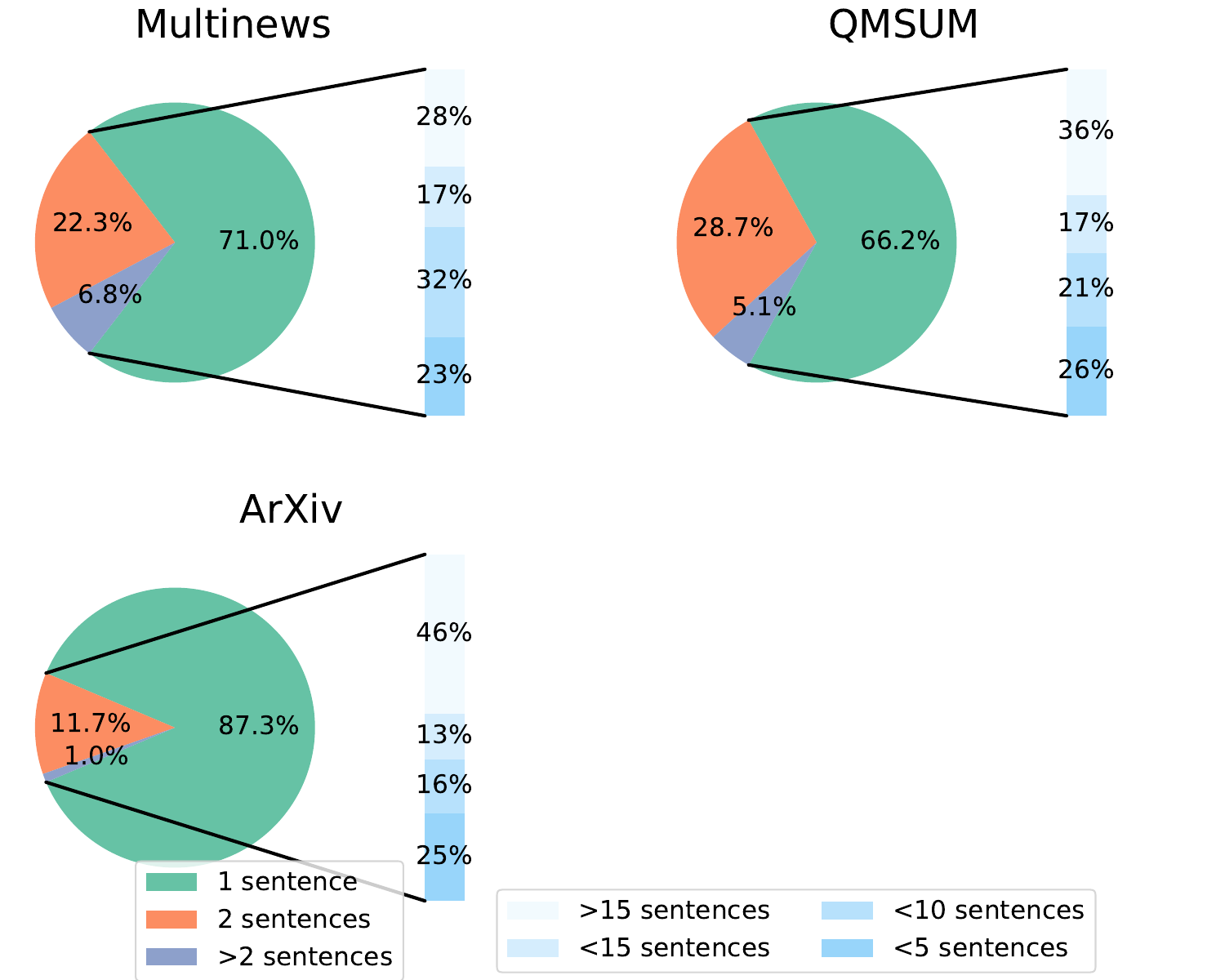}
    %\vspace{-0.5em}
    \caption{Statistics for the number of fused document sentences (the pie charts) and their distances (the blue vertical bars) on qmsum, multinews, and arxiv (DiverSumm).
    }
    \label{fig:sentence_fusion_appendix}
\end{figure}

\begin{figure*}[t]
    \includegraphics[width=1\linewidth]{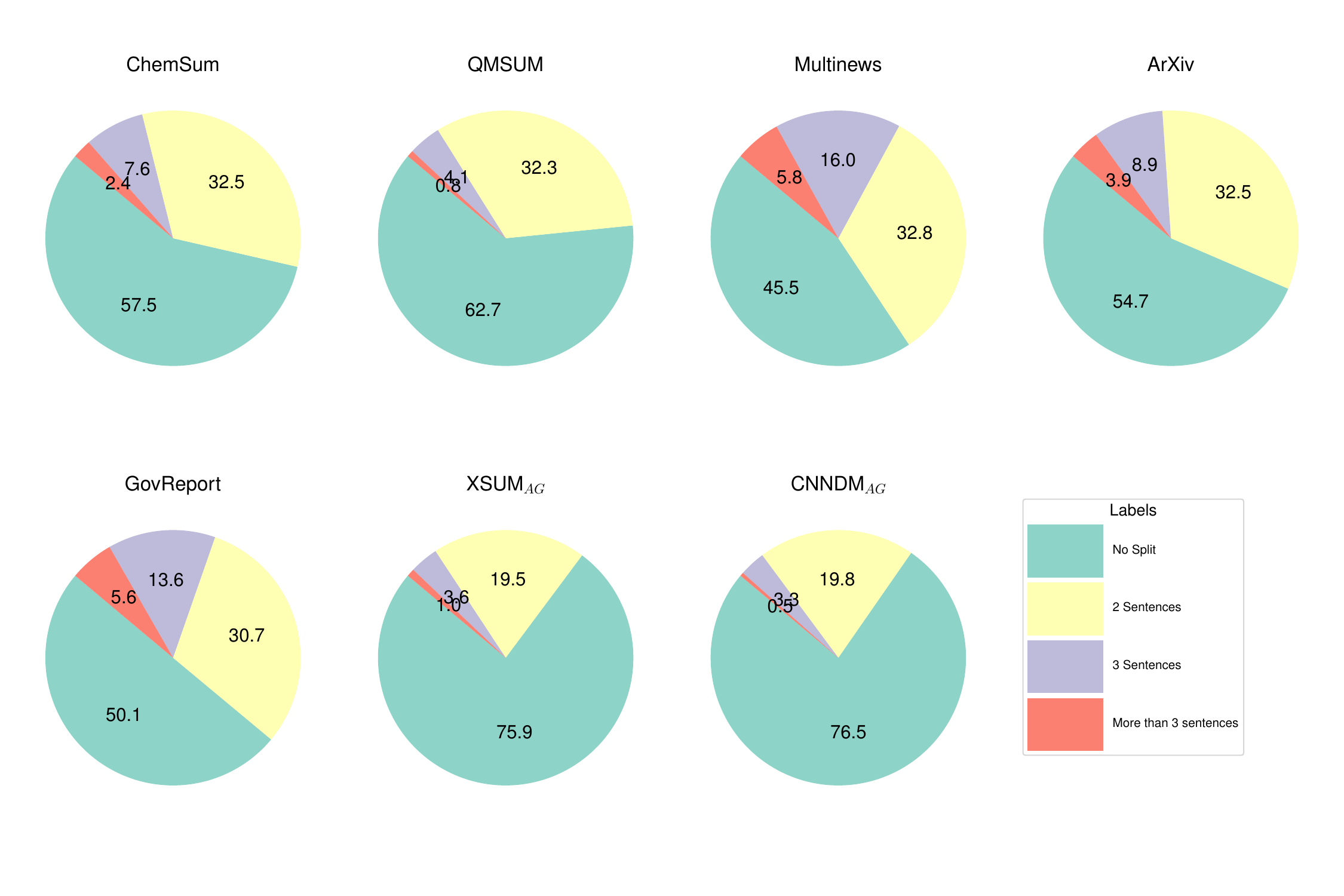}
    \caption{Distribution of number of splits occurring in summary sentences.
    }
    \label{fig:num_split}
\end{figure*}

\begin{table*}[t]
\centering
\small
\def\arraystretch{1.5}
\begin{tabular}{ll}
\thickhline
Sentence & Sub-sentences\\
\hline
\makecell{
Heritage auctions offered the gray jacket featuring\\ a black zigzag applique}
&
\makecell{
Heritage auctions offered the gray jacket.\\
The gray jacket featured a black zigzag applique.
}
\\
\hline
\makecell{
S.t. Mirren have signed striker Jeremy Clarkson \\on a season-long loan from Dundee. 
}
&
\makecell{
S.t. Mirren have signed striker Jeremy Clarkson.\\
The striker is on a season-long loan from Dundee.
}
\\
\hline
Change is a problem for many disabled people. &
Change is a problem for many disabled people.\\
\hline
\thickhline
\end{tabular}
\caption{\label{tab:subsentence}Examples of original sentences and their rewritten sentences for sub-sentence reasoning.
}
\end{table*}

\paragraph{Annotated Sets in AggreFact}

FactCC by \citet{kryscinski-etal-2020-evaluating} and SummEval \citet{fabbri-etal-2021-summeval}  are annotated at summary level. FactCC uses a binary consistency label (consistent/inconsistent). SummEval uses a 5-point Likert scale where only a score of 5 is treated as correct while the rest are considered incorrect.  

The annotation of Wang’20 \citet{wang-etal-2020-asking} and FRANK \citet{DBLP:conf/naacl/PagnoniBT21} operates at sentence level. Wang’20 employs a binary consistency label (consistent/inconsistent). 
A summary is labelled as faithful (consistent) if all of its sentences are labelled as consistent. 
The annotation scheme in FRANK highlights faithfulness error types (see Error Types in Section~\ref{sec:faithfulness_sum_tasks}) in summary sentences. 
Summaries in FRANK are considered to be faithful if none of their sentences are annotated with errors.

Polytope \cite{huang-etal-2020-achieved}, XSumFaith \citet{maynez-etal-2020-faithfulness} and Goyal’21 \citet{goyal-durrett-2021-annotating} are annotated at span level. Polytope identifies various error types such as addition, omission, and intrinsic inaccuracies. The annotation of XSumFaith revolves around error types like intrinsic and extrinsic. Goyal’21 classifies the error types into {intrinsic, extrinsic} × {entity, event, noun phrase}. 
Summaries devoid of these errors are marked as faithful.

CLIFF \citet{cao-wang-2021-cliff} is annotated at word level and its annotation scheme accounts for instinct/extrinsic hallucinations and lack of world knowledge. Cao’22 by \citet{cao-etal-2022-hallucinated} annotates entities and categorizes incorrect entities into factual/non-factual/instinct hallucinations. 
Summaries devoid of these errors are marked as faithful.

For details on the annotation process, we refer the reader to Aggrefact \cite{aggrefact}.

\paragraph{License}
No license is found for AggreFact, GovReport and ChemSumm.
ArXiv is under Apache-2.0 license.
QMSUM and MultiNews are under MIT License.  We ensure that the data was used solely for academic purposes, which aligns with the intended use of these datasets.
For data safety, content filtering was conducted when the creators built the original datasets. It is not avoidable that some documents can contain uncomfortable content, including news coverage of crimes and wars. For the model-generated summaries annotated with human judgements collected from \cite{chen-etal-2023-unisumm,koh-etal-2022-far,adams-etal-2023-desired} to create DiverSumm, we download some sets from their corresponding online download link and make others directly facilitated by the authors available in our github.\footnote{\url{https://huggingface.co/datasets/griffin/ChemSum}, \href{https://github.com/huankoh/How-Far-are-We-from-Robust-Long-Abstractive-Summarization}{https://github.com/huankoh/How-Far-are-We-from-Robust-Long-Abstractive-Summarization}.} We obtained permission from the authors for their use and encourage citation of the sets' corresponding work upon their future use within DiverSumm. We use these annotated sets only for research purposes.

\begin{table*}[t]
\centering
\scriptsize
\setlength{\tabcolsep}{2pt}
\def\arraystretch{1.5}
\begin{tabular}{lccccccc}
\thickhline
Models & XSM$_\textsc{ag}$ & CND$_\textsc{ag}$ & CSM & QMS & AXV & GOV &  MNW   \\

\ours &  2.66$\pm$1.67 & 1.79$\pm$1.04 & 2.50 $\pm$4.76 & 2.55 $\pm$1.48&  4.89 $\pm$9.25 & 2.11$\pm$1.21&1.98 $\pm$1.25 \\

\oursub & 2.40$\pm$1.57 & 1.22$\pm$1.75 & 2.12 $\pm$3.79 & 2.41 $\pm$1.43 &  4.09 $\pm$8.26 & 1.92 $\pm$1.11& 1.76 $\pm$1.09  \\

\thickhline
\end{tabular}
\caption{\label{tab:kvalue} Average number of retrieved document sentences and standard deviation for \ours~  and \oursub~ on AggreFact and DiverSumm. }

\end{table*}

\begin{table}[t]
\centering
\scriptsize
\setlength{\tabcolsep}{2pt}
\def\arraystretch{1.5}
\begin{tabular}{lcc}
\thickhline
Approach & AUC & Nb. calls to NLI \\
\fulldoc &58.74&1  \\
\textsc{Summac}$_\textsc{conv}$ & 60.54 & M+C \\
\textsc{Summac}$_\textsc{zs}$ & 63.78 & M \\
\sentli & 62.56 & M+1\\ 
\ours-$k$ & 65.01 & 2M+1   \\
\ours &  65.84  & 2M+$k_{avg}$+1 \\

\thickhline
\end{tabular}
\caption{\label{tab:tradeoff} Performance / Computation trade-off. We report the AUC versus the number of calls to the NLI model. M is the number of document sentences. $k_{avg}$ is the expected number of retrieved document sentences which can entail the summary sentence. C is the call to the convolution layer. 
}
\end{table}

\section{Training Configurations}
\label{appendix:training}

\paragraph{Models}

We use the publicly-available 
\url{https://huggingface.co/tals/albert-xlarge-vitaminc-mnli} NLI model. We use the tokenizer from Stanza \cite{qi2020stanza}. 

Originally \textsc{Summac}$_\textsc{ZS}$ uses the combination of entailment - contradiction which was found to perform better \cite{laban-etal-2022-summac}. However, we find that in both AggreFact and DiverSumm, by taking only the entailment score \textsc{Summac}$_\textsc{ZS}$ obtains a much better performance. Thus, we only use entailment scores. In addition, the implementation of \textsc{Summac} ignores those document sentences with less than 10 tokens and only considers the first 100 sentences of the document. We remove such constraints for a fair comparison. In addition, \textsc{Summac} obtains better performance without these constraints.

\begin{table*}[t]
\centering
\footnotesize
\setlength{\tabcolsep}{2pt}
\def\arraystretch{1.2}
\begin{tabular}{lcacacacacacacaca}
\thickhline
 & \multicolumn{2}{c}{XSM$_\textsc{ag}$} & \multicolumn{2}{c}{CND$_\textsc{ag}$} & \multicolumn{2}{c}{CSM} & \multicolumn{2}{c}{QMS} & \multicolumn{2}{c}{AXV} & \multicolumn{2}{c}{GOV} &   \multicolumn{2}{c}{MNW} & \multicolumn{2}{c}{AVG}  \\
\hline
\textsc{context} & \textsc{sent} & \multicolumn{1}{c}{\textsc{sub}} & \textsc{sent} & \multicolumn{1}{c}{\textsc{sub}} & \textsc{sent} & \multicolumn{1}{c}{\textsc{sub}} & \textsc{sent} & \multicolumn{1}{c}{\textsc{sub}} & \textsc{sent} & \multicolumn{1}{c}{\textsc{sub}} & \textsc{sent} & \multicolumn{1}{c}{\textsc{sub}} & \textsc{sent} & \multicolumn{1}{c}{\textsc{sub}} & \textsc{sent} & \multicolumn{1}{c}{\textsc{sub}} \\
\hline
\fulldoc & 72.77 & \textbf{\textcolor{olive}{73.63}} & 64.40 & 63.68 & 50.15 &  \textbf{\textcolor{olive}{58.72}} & 37.12 &  \textbf{\textcolor{olive}{39.76}} & 62.78 &  62.46 & 79.19 & 77.69 & 44.76 & \textbf{\textcolor{olive}{46.72}} & 58.74 & \textbf{\textcolor{olive}{60.38}}\\

\textsc{Summac}$_\textsc{conv}$  & 67.76&65.77&	72.14&	70.84&53.14&51.10&	51.13 & \textbf{\textcolor{olive}{54.42}} &	61.22&44.26&	65.34 & \textbf{\textcolor{olive}{81.58}}&	53.05 & \textbf{\textcolor{olive}{56.27}} & 60.54 & \textbf{\textcolor{olive}{60.61}}  \\
							
\textsc{Summac}$_\textsc{zs}$ & 70.29&66.67&	74.54& \textbf{\textcolor{olive}{74.98}} & 54.41& \textbf{\textcolor{olive}{57.32}}&	48.21&\textbf{\textcolor{olive}{51.42}}&69.44&67.26 &	79.37&\textbf{\textcolor{olive}{81.09}}&	50.17 &\textbf{\textcolor{olive}{54.20}}&	63.78 &\textbf{\textcolor{olive}{64.71}}\\

\sentli  & \textbf{73.61} & 71.45 & 75.83 & 74.66 & 50.13 & \textbf{\textcolor{olive}{55.69}} & 47.56 & \textbf{\textcolor{olive}{51.88}} & 64.49 & \textbf{\textcolor{olive}{76.35}} & 79.68 & 77.65 & 46.61 & 43.61 & 62.56 & \textbf{\textcolor{olive}{64.47}} \\

%\rowcolor{Gray}
\ours & 73.42 & \multicolumn{1}{c}{73.21} & \textbf{76.21} & \multicolumn{1}{c}{73.34} & \multicolumn{1}{c}{54.11} &  \multicolumn{1}{c}{\textbf{59.26}} & \multicolumn{1}{c}{52.16} & \multicolumn{1}{c}{\textbf{53.20}} & \multicolumn{1}{c}{71.38} & \multicolumn{1}{c}{\textbf{73.89}} & \multicolumn{1}{c}{\textbf{80.45}} & \multicolumn{1}{c}{80.05} & \textbf{53.16} & \multicolumn{1}{c}{49.37} & \textbf{\underline{65.84}} & \multicolumn{1}{c}{\textbf{\underline{66.05}}} \\

\hline

\thickhline
\end{tabular}
\vspace{-0.5em}
\caption{\label{tab:res_aggrefact_and_diversumm_long}
Results for all summarisation tasks in AggreFact and DiverSumm combined with summary sentence splitting (SUB column).  
For AggreFact, we report the average results for XSum (XSM; 5 datasets) and CNN/DM (CND; 7 datasets), respectively; dataset-level performance can be found in Appendix \ref{appendix:more_results}. 
CSM, MNW, QMS, AXV, and GOV refer to ChemSumm, MultiNews, QMSUM, ArXiv, and GovReport respectively. 
We highlight \textbf{highest} scores and scores \underline{significantly different} from \fulldoc, all \textsc{SummaC} variants and \sentli~models (at $p<.05$, using pairwise t-test).  We additionally highlight in \textbf{\textcolor{olive}{olive}} improved scores for existing approaches when combined with summary sentence splitting.
}

\end{table*}

\begin{figure}[t]
    \centering
    \vspace{-1.0em}
    \includegraphics[width=7.0cm]{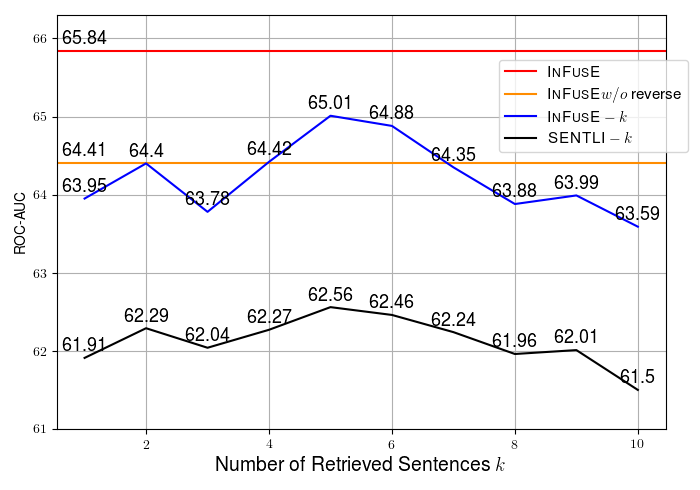}
    \vspace{-.8em}
    \caption{\label{fig:premise_size}Performance over retrieval size $k$. We report the average ROC-AUC on AggreFact and DiverSumm. 
    }
\end{figure}

\begin{figure}[h]
    \centering
    \begin{subfigure}{.30\textwidth}
    \centering
    \includegraphics[width=1\linewidth]{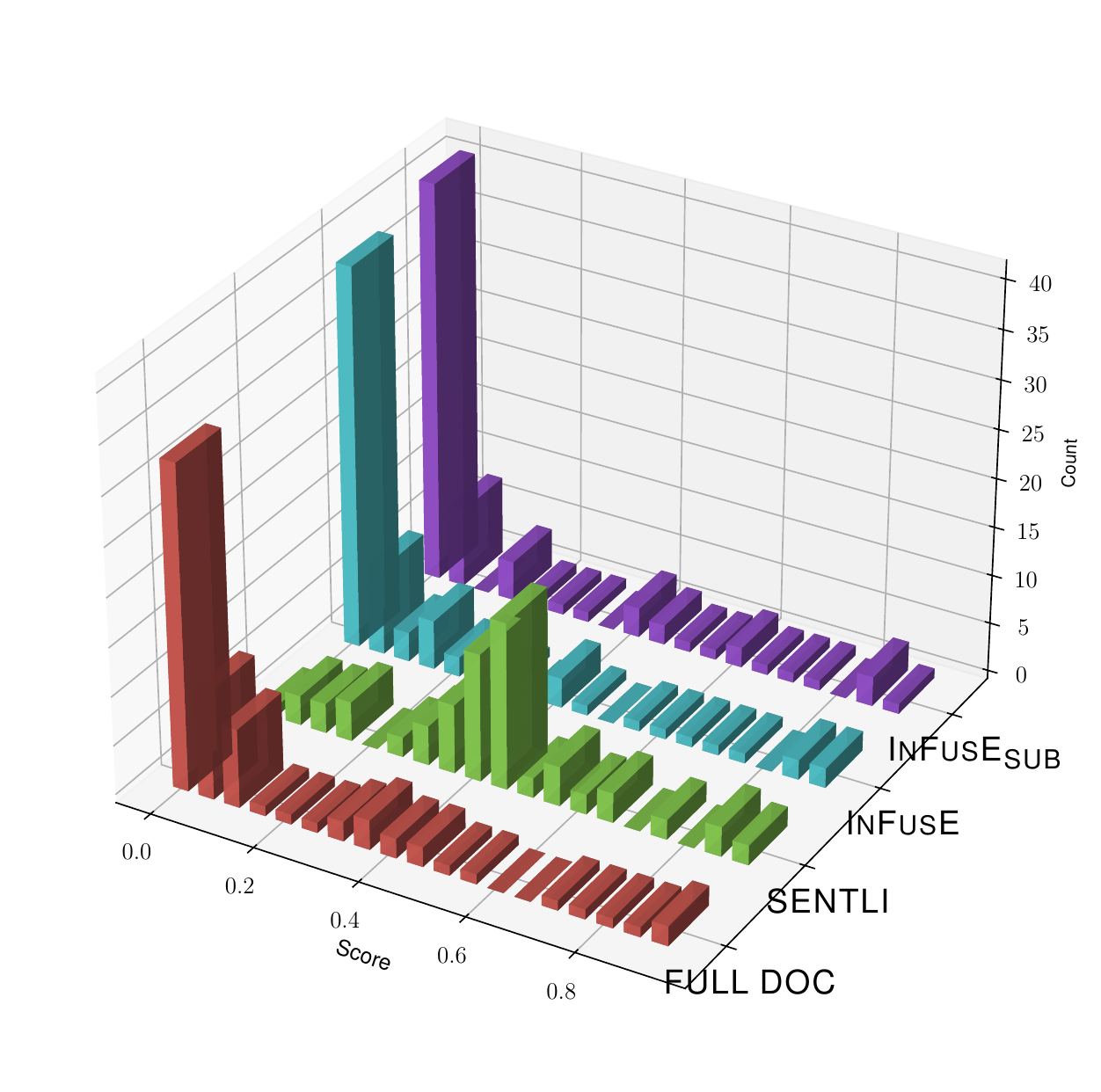}
    \vspace{-0.8cm}
    \caption{CircE}
    \label{fig:sfig2}
    \end{subfigure}\\
    \begin{subfigure}{.30\textwidth}
    \centering
    \includegraphics[width=1\linewidth]{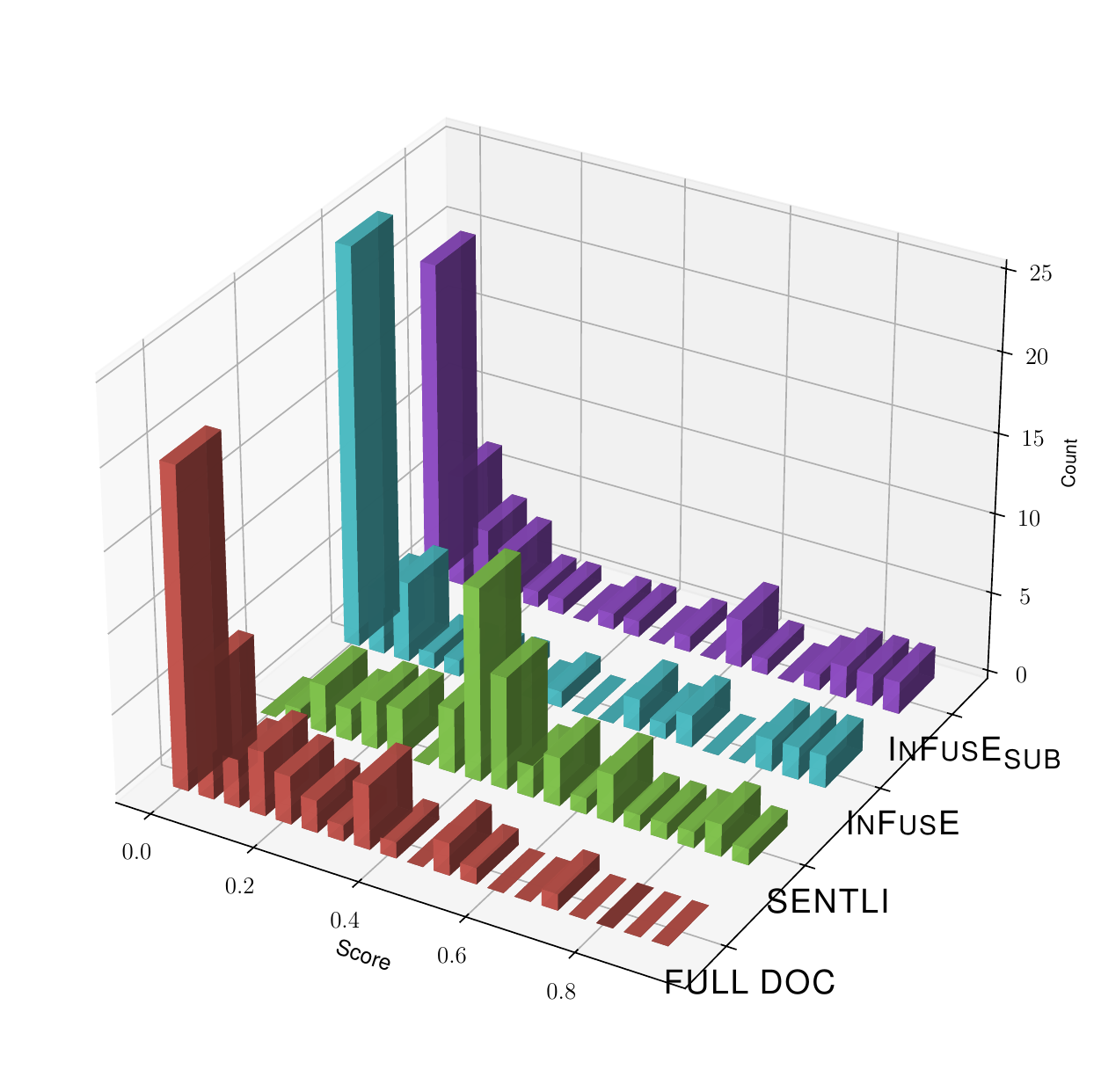}
    \vspace{-0.8cm}
    \caption{PredE}
    \label{fig:sfig4}
    \end{subfigure}
    \caption{Distribution of entailment scores on correct and different error types for arXiv and GovReport from DiverSumm. The x-axe corresponds to the NLI-based approach. That is, \fulldoc~ in red, \sentli~ in green, \ours~ in cyan, and \oursub~ in purple. The y-axe corresponds to the entailment scores (i.e., values ranging in [0,1]), and the z-axe corresponds to the count of instances.}
    \label{fig:error_types_ent_dist_add}
\end{figure}

\paragraph{Sentence Splitting}
\citet{wice} propose a dataset, namely WiCE, including original claim sentences paired with their decomposition (split) into more than one sentence generated by GPT-3 \cite{NEURIPS2020_1457c0d6}. We leverage such parallel data to train a sentence splitting model for sub-sentence reasoning based on T5-large \cite{2020t5}.
We fine-tune T5-large for 5 epochs with a batch size of 32 and a learning rate 5e-4. We force the length of the output to be within $[3,128]$.
We show a few sentence splitting examples in Table \ref{tab:subsentence}. Figure~\ref{fig:num_split} shows the distribution of the number of splits that summary sentences had. 
We train the model on an A6000 GPU and each epoch costs 90 seconds. The inference time is around 8 sample per second.

\section{Summary Sentence Splitting is Beneficial for All Approaches}
\label{app:sub_approaches}

Table~\ref{tab:res_aggrefact_and_diversumm_long} shows additional results when we combine the proposed summary sentence splitting step with the different approaches to build a premise. We can see that sub-sentence (SUB column in Table~\ref{tab:res_aggrefact_and_diversumm_long}) brings improvements across all of them (as discussed before with the exception of CNN/DM).
Sub-sentence evaluation brings improvements for sentence-level premises such as SUMMAC in particular for the version that relies on a convolutional neural network trained to map the distribution of entailment scores to correct/incorrect judgements. After splitting there is less content fusion from document sentences and more feasible to judge entailment with one document sentence.

For ArXiv and CSM context-level works better indicating that neither one sentence nor the entire document provide adequate context even after summary sentence splitting.
For XSum, the most abstractive dataset with short input documents, the document-level (\fulldoc) and context-level (\ours~ and \sentli) premises work well. For this dataset sentence-level approaches (\summac) even with sentence splitting are not enough.
Overall, \ours~ and \oursub~ perform the best, this shows that the variable context allows to account for different levels of document sentence fusion.

\section{Dataset-Level Performance on AggreFact}
\label{appendix:more_results}

\begin{table*}[t]
\centering
\small
\setlength{\tabcolsep}{1.5pt}
\def\arraystretch{1.5}
\begin{tabular}{lcccccc|cccccccc}
\thickhline
\multirow{2}{*}{Models} & \multicolumn{6}{c|}{XSum Test} & \multicolumn{8}{c}{CNN/DM Test}      \\
 ~& \multicolumn{1}{l}{Wang'20} & \multicolumn{1}{l}{Cao'22} & \multicolumn{1}{l}{XSF} & \multicolumn{1}{l}{Goyal'21} & \multicolumn{1}{l}{CLF} & \multicolumn{1}{l}{\textbf{AVG}} & \multicolumn{1}{l}{FCC} & \multicolumn{1}{l}{Wang'20} & \multicolumn{1}{l}{SEV} & \multicolumn{1}{l}{PTP} & \multicolumn{1}{l}{FRK} & \multicolumn{1}{l}{Goyal’21} & \multicolumn{1}{l}{CLF} & \multicolumn{1}{l}{\textbf{AVG}} \\
 
\hline

\fulldoc  & 64.62&	\textbf{71.50}&	\textbf{75.76}&	74.70&	77.34&	72.78
& 75.63&	\textbf{84.09}&	74.07&	76.69&	65.26&	\hspace{0.5em}4.17&	70.90&	64.40
\\

\textsc{Summac}$_\textsc{Conv}$  &

69.59&
69.71&
70.03&
56.40&
73.09&
67.76&

92.22&
76.67&
85.48&
81.67&
76.72&
\textbf{25.00}&
67.20&
72.14

\\
\textsc{Summac}$_\textsc{ZS}$   & 
73.77&
67.27&
72.73&
61.58&
76.11&
70.29& 
93.72&
80.94&
87.57&
88.57&
77.22&
\textbf{25.00}&
68.81&
74.54

\\
\sentli  &72.80&	70.57&	69.46&	\textbf{74.88} &	80.34&	\textbf{73.61} &
92.26&	80.04&	87.75&	92.82&	\textbf{79.92}&	20.83&	\textbf{77.23}&75.83

\\

\ours  & \textbf{\underline{76.41}}  & 67.73 & 74.01    & 71.43   & 77.51      & 73.42  & 
\textbf{\underline{94.99}} & 80.21   & \textbf{88.65}   & \textbf{92.84}   & 79.48       & 20.83   & 76.45       & \textbf{76.21}  \\

\ours$_\textsc{sub}$ & 73.76&	69.92	&74.69	&66.34&	\textbf{\underline{81.36}}& 73.21&
92.73&78.66&87.76&83.68&77.76&16.67&72.82 &72.87\\

\thickhline
\end{tabular}
\caption{\label{tab:detailed_aggrefact}Dataset-level performance on AggreFact. 
For XSum Test, XSF and CLF refer to XSumFaith and CLIFF, respectively. 
PTP, FCC, SEV and FRK refer to Polytope, FactCC, SummEval and FRANK, respectively. 
We highlight \textbf{highest} scores and scores \underline{significantly different} from \fulldoc, all \textsc{SummaC} and \sentli~models (at $p<.05$, using pairwise t-test).
}
\end{table*}

We show detailed results for AggreFact in Table \ref{tab:detailed_aggrefact}. Statistical significance of \ours~ w.r.t. to the other best performing approaches are computed as described in Section~\ref{sec:overall_auc}. Overall, there is no significant difference among \ours~, \sentli, and \fulldoc~ on XSum$_{AG}$ and CNNDM$_{AG}$. Interestingly, the models exhibit different performance within subsets of the tasks. \ours~is significantly better on Wand'20, CLIFF, and FactCC.

\section{Performance per Premise Sizes}
\label{app:performance_on_k}

Figure~\ref{fig:premise_size} shows the evaluation performance (ROC-AUC) for different premise sizes $k$ (i.e., number of document sentences). It includes \sentli~, a variant of \ours~with a fixed retrieval size (\ours-$k$), and \ours~without reverse reasoning. 
As can be seen, reversed reasoning helps to produce better entailment judgements as there is a performance degradation when we remove it from \ours. Incremental reasoning allows \ours~to determine \textit{when to stop} automatically, removing the requirement of additional data for optimizing the retrieval size $k$ which has a substantial impact on model performance.

Table~\ref{tab:kvalue} shows the average premise size, in number of document sentences, at which \ours~ and \oursub~ work. We can see that there is considerable variability in the number of retrieved sentences within and across tasks. This further supports the difference in performance between \ours~ and \ours-$k$.

\section{Performance per Error Types}
\label{app:analysis_error_type}

Figure~\ref{fig:error_types_ent_dist_add} shows two additional graphs for CircE and PredE error types. Similarly to EntE (Section~\ref{sec:performance_error_types}), \ours~ and \ours~ perform better than \sentli~ which assigns scores mainly in the [0.4, 0.6] interval. \ours~ performs better than \oursub. We show examples of cases correctly ($\sim$ 0) and incorrectly ($\sim$ 1) judged by \ours~ in Table~\ref{tab:high_low_error_types}.

\section{Case Studies}
\label{appendix:analysis}

\paragraph{Sentence Fusion}

To illustrate how sentence fusion renders difficult the assessment of entailment by current sentence-level NLI models, we provide two representative examples from the faithfulness evaluation benchmarks in Table~\ref{tab:error2}.

\begin{table*}[h]
\renewcommand{\arraystretch}{1.5}
    \centering
    \small
    \begin{tabular}{c|p{10cm}|c|c}
    \thickhline
    && DS $\models$ MSS  & MSS $\models$ DS\\
    \thickhline
     D & Sao Paulo, Brazil (CNN)Brazilian supermodel Gisele Bundchen sashayed down the catwalk at Sao Paulo Fashion Week on Wednesday night in an emotional farewell to the runway.& .003 & .003\\
          
&Bundchen announced over the weekend that she would be retiring from the catwalk, though not the fashion industry. & .004 & .003\\
&The 34-year-old, who is married to New England Patriots quarterback Tom Brady and has two children, has said she wants to spend more time with her family.& .001 & .001\\
&On Wednesday night, Brady had a front-row seat at what was hailed as a historic moment in Brazil's fashion world.& .006 & .003\\
&\textcolor{violet}{Bundchen wrote about her fashion career on her Instagram account}: "I am grateful that at 14, I was given the opportunity to start this journey.& .996 & .002 \\
&Today after 20 years in the industry, it is a privilege to be doing my last fashion show by choice and yet still be working in other facets of the business." & .002 & .001\\

     \hline
     MSS & 
     \multicolumn{3}{p{13cm}}{\textcolor{violet}{bundchen wrote about her fashion career on her instagram account.}} \\

      \thickhline   
D  &  \textcolor{violet}{David Lipton, second in command at the IMF}, outlined some of these risks in a speech to the National Association for Business Economics in Washington on Tuesday.  & .018 & .001\\

&"The IMF's latest reading of the global economy shows once again a weakening baseline," he said.& .103 & .006\\

&\textcolor{cyan}{"We are clearly at a delicate juncture."} & .020 & .212\\

&The comments come after weaker-than-expected trade figures from China showing that exports plunged by a quarter from a year ago.& .004 & .001\\

&The IMF has already said \textcolor[RGB]{0,176,80}{it is likely it will downgrade its current forecast of 3.4\% for global growth} when it next releases its economic predictions in April.& .050 & .020 \\

&The dismal picture is one that has on-going ramifications for businesses and industries that bet on China's growth story.& .002 & .003\\

&Read more from Karishma: & .002 & .004\\

&Why a story about bulk shipping matters.& .002 & .019\\
    \hline
        MSS & \multicolumn{3}{p{13cm}}{\textcolor{violet}{The head of the International Monetary Fund (IMF)} has warned that \textcolor{cyan}{the global economy is "at a delicate juncture" }and that \textcolor[RGB]{0,176,80}{the outlook for global growth is "deteriorating".} } \\

    \thickhline
     
    \end{tabular}
    \caption{\label{tab:effects_of_reverse} We show input Document (D), Model-generated Summary Sentence (MSS), and DS $\models$ MSS (Document Sentence -DS- to summary sentence reasoning) and MSS $\models$ DS (reversed reasoning) scores. We highlight content segments in summary sentences and their corresponding document evidence in \textcolor{violet}{violet}, \textcolor{cyan}{cyan} and \textcolor[RGB]{0,176,80}{green}. The example in the top part is from FactCC \cite{kryscinski-etal-2020-evaluating} in {CNNDM$_\textsc{ag}$} and the second is from CAO'22 \cite{cao-etal-2022-hallucinated} in {XSUM$_\textsc{ag}$}. Both labelled as faithful (correct) summaries.
    }
    \label{tab:error2}
\end{table*}

\begin{figure}[h]
    \centering
    \vspace{-1.0em}
    \includegraphics[width=7.0cm]{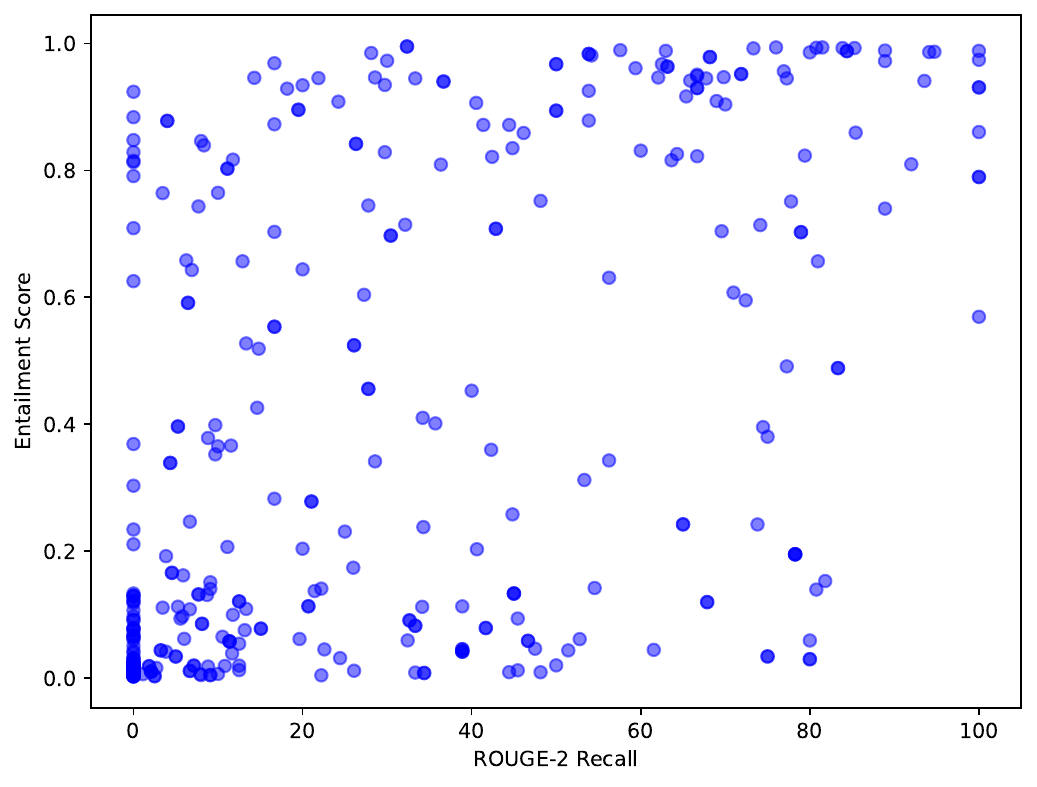}
    \vspace{-.8em}
    \caption{\label{fig:ent_bias_rouge}
    ROUGE-2 Recall versus Entailment Score on summary sentences labelled as unfaithful from the ArXiv and GovReport datasets.
    }
\end{figure}

\begin{figure}[h]
    \centering
    \vspace{-1.0em}
    \includegraphics[width=7.0cm]{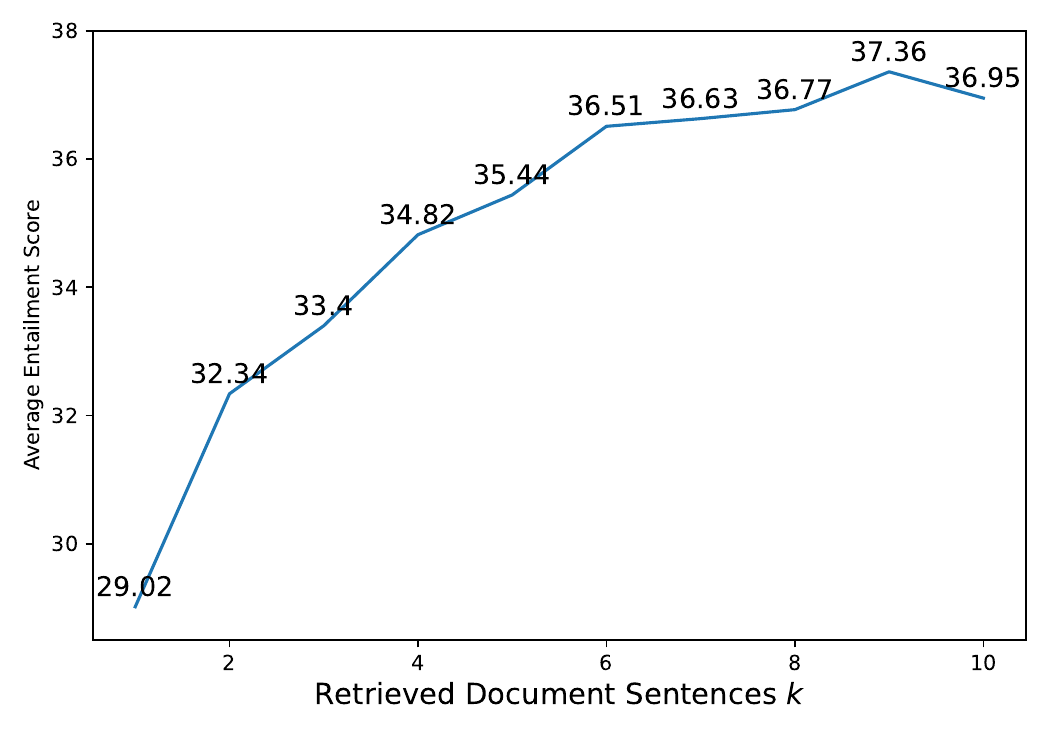}
    \vspace{-.8em}
    \caption{\label{fig:ent_bias_len}
    Average entailment score for summary sentences labelled as unfaithful from the
ArXiv and GovReport datasets. Premise size, in number of retrieved document sentences, ranges from 1 to 10. 
    }
\end{figure}

    The first example, taken from FactCC \cite{kryscinski-etal-2020-evaluating}, shows a  summary that is simply a short version of one document sentence. In these cases, a sentence-level NLI evaluator \cite{laban-etal-2022-summac,nie-etal-2020-adversarial}
    would capture the relation and assign a high entailment score. 

    In contrast, the second example from CAO'22 \cite{cao-etal-2022-hallucinated} is more complex: the content conveyed in the summary sentence fuses content included in multiple document sentences. In this situation, three possibilities arise. First, if the summary sentence is more informative than a document sentence and the part that overlaps is a paraphrase, it can be captured by applying NLI in reversed direction (i.e., summary-to-document, MSS $\models$ DS column in Table~\ref{tab:effects_of_reverse}). Examples of this scenario are text segments highlighted in \textcolor{cyan}{cyan}. Second, if none of the inference directions (neither document-to-summary, DS $\models$ MSS column in Table~\ref{tab:effects_of_reverse}, nor MSS $\models$ DS) achieve a high entailment score individually, the combined score may still be relatively high allowing the bidirectional method in \ours~ to capture such cases as illustrated by the example in  \textcolor[RGB]{0,176,80}{green}. Third, a content unit in a complex and informative summary sentence is entailed by a content unit in a complex document sentence they only overlap on this content unit.
    It is possible that the method will fail in these cases, as the sentence segments in \textcolor{violet}{violet} illustrate.

\paragraph{High Reversed Reasoning Scores}

Table~\ref{tab:reverse_examples} shows examples of document and summary sentence inference applied in both the standard and reversed direction (lines 2 and 3 in Algorithm~\ref{alg:infuse}).
These examples are taken from summaries annotated as (correct) faithful.
In particular, these show cases where the reversed direction yields high entailment scores.
These are cases where the summary sentence is providing more details due to sentence fusion. For instance, in the third example, the summary sentence is adding extra information (taken from other document sentences) about \textit{Paulo Duarte} being \textit{Burkina Faso's coach}.
Note that in some cases sentences contain pronouns and thus they should not lead to high entailment scores because the referent is unknown \cite{delmonte-etal-2007-entailment}. However, the NLI model is biased because of the premise-hypothesis length and token overlap \cite{mckenna2023sources,verma-etal-2023-evaluating}.

\begin{table*}[h]
\renewcommand{\arraystretch}{1.5}
    \centering
    \footnotesize
    \begin{tabular}{p{5.5cm}|p{5.5cm}|c|c}
    \thickhline
    DS & MSS & DS $\models$ MSS & MSS $\models$ DS \\
    \thickhline

 He resigned from his post in order to make this appearance.
 & A police chief resigned from his post to appear on bbc question time.
 & .003 & .938 \\

 We will be making no appeal.
 & Wigan warriors will not appeal against the eight-game ban given to ben flower for punching st helens prop lance hohaia.
 & .004 & .930 \\

"I am confident they can recover in time," Duarte insisted.
& Burkina faso coach paulo duarte says he is confident his players will be fit for next month's africa cup of nations.
& .013 & .388 \\

In a statement the company said the blaze had affected an estimated 1,000-2,000 tonnes of recycled wood chip.
& Firefighters are continuing to tackle a blaze at a wood chip recycling plant in Bridgend county which has destroyed up to 2,000 tonnes of wood chip.
& .019 & .067 \\

Decisions about which people, groups, or events to memorialize are made by many different entities, including Congress, federal agencies, state and local governments, and private citizens, among others. 
& Decisions about which people, groups (or events), and which places to memorialize, are made by many different entities, including Congress, federal agencies, state and local governments, and private citizens, among others.
& .091 & .980 \\

NOAA has defined natural infrastructure and nature-based infrastructure in NOAA Administrative Order (NAO) 216-117: NOAA National Habitat Policy.
& NOAA's National Habitat Policy (NAO 216-117) directs the agency to protect, maintain, and restore ocean, coastal, and Great Lakes ecosystems by "applying natural and natural infrastructure," among other activities.
& .007 & .685 \\

This report considers the extent of federal involvement in memorials located outside the District of Columbia (Washington, DC). 
& This report considers the extent of federal involvement in national memorials located outside the District of Columbia (Washington, DC).
& .166 & .981 \\

In the United States, there are hundreds, and possibly thousands, of memorials to various individuals, groups, and events. 
& In the United States, there are hundreds, and possibly thousands, of memorials to various individuals, group, and historical events.
& .224 & .989 \\

    \thickhline
     
    \end{tabular}
    \caption{Examples of reversed reasoning with high entailment scores. Document Sentence (DS), Model-generated Summmary Sentence (MSS), document to summary entailment (DS $\models$ MSS), and reverse direction (MSS $\models$ DS). All examples are from summaries in the DiverSumm benchmark labelled as faithful (correct).}
    \label{tab:reverse_examples}
\end{table*}

%https://academic.oup.com/book/27086/chapter/196432173

\paragraph{$\sim 0$ and $\sim1$ Entailment Scores on Different Error Types}

Table~\ref{tab:high_low_error_types} shows examples of \ours~ working on FRANK, ArXiv, and GovReport (Section~\ref{sec:performance_error_types}). The top part of the table includes cases where \ours~ successfully assigns close to zero scores to unfaithful cases per error type and the bottom part illustrates those scenarios where it fails to identify the error. On manual inspection, we find that in many cases these failures are related to high lexical overlap and premise-hypothesis length bias \cite{mckenna2023sources,verma-etal-2023-evaluating}. 
Figure~\ref{fig:ent_bias_rouge} and Figure~\ref{fig:ent_bias_len} show this trend for all unfaithful sentences in the ArXiv and GovReport subsets. We observe a similar trend in all datasets in AggreFact and DiverSumm but only these two datasets have sentence level annotation. 

In Figure~\ref{fig:ent_bias_rouge}, we analyse entailment scores for premise-hypothesis pairs in relation to their lexical overlap.\footnote{Note that by premise we mean the premise selected by \ours.} We compute lexical overlap as ROUGE-2 Recall in order to capture phrase information. As can be seen, on the left-bottom corner, a high percentage of pairs with low ROUGE-2 Recall obtain a low entailment score. Another cluster of pairs is on the right-top corner where pairs with high lexical overlap get high entailment scores. This behaviour of NLI models will undermine evaluation of summary faithfulness when summaries are abstract or have a high token overlap but differ in few words that change the content conveyed in the input document (e.g., negation). Figure~\ref{fig:ent_bias_len} shows average entailment scores in relation to premise size. That is, we compute average entailment scores for premise-hypothesis pairs setting the premise to the top $k$ ranked document sentences; $k$ takes values from 1 to 10. We can see that longer premises obtain higher entailment scores despite the fact that they include document sentences further below in the rank.

\begin{table*}[h]
\renewcommand{\arraystretch}{1.5}
    \centering
    \scriptsize
    \begin{tabular}{p{6cm}|p{5cm}|p{0.6cm}}
    \thickhline
   DRS & MSS & Error Type \\
    \thickhline
    \multicolumn{2}{c}{Entailment Scores $\sim 0$}\\
\hline
  Costs for Group B benefits and administration are financed by the one-time appropriation of \$4.6 billion provided in the Zadroga Reauthorization Act of 2015. 
  & Costs for Group B benefits and administrative expenses were financed by a one-time appropriation of \textcolor{red}{\$3}.
  & EntE\\

  Jan 2006 - Government proposes nuclear as part of future energy mixMar 2013 - Construction of Hinkley Point approvedOct 2013 - UK government agrees £92.50 per megawatt-hour will be paid for electricity produced at the Somerset site - around double the current market rate at the timeOct 2015 - EDF signs investment agreement with China General Nuclear Power Corporation (CGN)July 2016 - EDF board approves final investment decision, but the UK Government postpones a final decision on the project until autumn. 
  &The government has given the \textcolor{red}{go-ahead} for a new nuclear power plant at a former nuclear plant in somerset. 
  & PredE \\
  
  The VCF was reauthorized in 2015 and, if not reauthorized in the 116 th Congress, will sunset on December 18, 2020.
  &The MTF was reauthorized in 2015 and, if not reauthorized, the current iteration will sunset \textcolor{red}{on June 18, 2017}.
  & CircE \\
  
  While men caregivers may face some of these risks, the effects of  caregiving for women are compounded by lower average lifetime  earnings and a longer life expectancy than men. As a result, women  caregivers are at an increased risk of outliving their savings.
  &Women caregivers were more likely than men caregivers to be employed \textcolor{red}{and to have higher levels of earnings}, but women caregivers were also more likely to work part-time \textcolor{red}{and have lower levels of employment} and have less income.
  & LinkE\\
  
  In our December 2018 report, we found that TSA provides pipeline  operators with voluntary security guidelines that operators can implement  to enhance the security of their pipeline facilities. 
  &The Transportation Security Administration (\textcolor{red}{TSAO}) provides pipeline operators with voluntary security guidelines that operators can implement to enhance the security of their pipeline facilities. 
  & CorefE\\
  
  Since fiscal year 2008, the United States has  allocated about \$3 billion for assistance for Mexico under the Mérida  Initiative. You asked us to review issues related to Mérida Initiative implementation  and objectives. 
  &Since fiscal year 2008, the United State has allocated about \$3 billion for assistance for Mexico under the \textcolor{red}{Civil Standards Initiative}. 
  & OutE\\

  In July 2016, OMB issued an updated Circular No. A-123, Management’s  Responsibility for Enterprise Risk Management and Internal Control,  which requires executive agencies to implement enterprise risk  management (ERM) in their management practices. Since the July 2016 update to OMB Circular No. A-123  required agencies to implement ERM, the Air Force has been leveraging  and relying on its existing risk management practices.
  &In July 2016, OMB issued an updated Circular No A. B, Management's Responsibility for Enterprise Risk Management and Internal Control, which requires executive \textcolor{red}{[incomplete sentence]}
  & GramE \\
  
  \hline

   % \thickhline
    % \multicolumn{2}{c}{Entailment Scores $\sim 0$}\\
\multicolumn{2}{c}{Entailment Scores $\sim 1$}\\
\hline
  Practitioners and decisionmakers have been using the term nature-based infrastructure and supporting nature-based infrastructure features since at least the late 2000s (although these types of features have been assigned various names over time) 
  &Practitioners and decisionmakers have been using the term \textcolor{red}{nature-by-nature-infrastructure} since at least the late 2000s, although these types have been assigned various names over time. & EntE \\
  
  Memorials with "medium" federal involvement typically either are located on federal land but do not receive federal funding, or are located on nonfederal land but receive assistance from a federal agency.
  &Memorials for purposes of "medium" involvement are either located on nonfederal land but \textcolor{red}{do not receive} federal funding, or are located in federal land but \textcolor{red}{receive} federal assistance from a federal agency.
  & PredE \\
  But he now faces at least a year at a militant rehabilitation centre in Kuwait, according to the terms of the release. The Kuwaiti government had pushed hard for the release of all Kuwaiti detainees at Guantanamo. 
  &A former guantanamo bay detainee has been released \textcolor{red}{from kuwait}. 
  & CircE \\

  The value of the 15 State projects in our  sample is about \$88 million, and the value of the five USAID projects in  our sample is about \$107 million. Because State/INL implemented about  90 percent of Mérida Initiative projects during this period, we chose a  larger State/INL sample than a USAID sample.
  & State/INL \textcolor{red}{and USAID} have implemented about 90 percent of MérIDA Initiative projects.
  & LinkE \\

  Administrators of the ACT test took the decision just hours before some 5,500 students were due to sit it. The other entrance exam - the SAT - was cancelled in South Korea in 2013 because some of the questions were leaked. 
  &A number of students have been barred from taking part in \textcolor{red}{a test test test in south korea}. 
  & CorefE\\

  But Prof Peter Godfrey-Smith said the unique study, based on 53 hours of footage and published on Friday in the journal Current Biology, provided a novel perspective on octopus behaviour."[An aggressive] octopus will turn very dark, stand in a way that accentuates its size and it will often seek to stand on a higher spot," Prof Godfrey-Smith, who co-authored the report, said.
  &\textcolor{red}{One of the world's most aggressive octopuses} appears to show signs of aggressive behaviour, a study suggests.
  & OutE \\

  No systematic law or set of regulations governs the establishment of memorials outside Washington, DC. 
  & No systematic law or set of regulations governs the establishment of \textcolor{red}{memorialses} outside Washington, D.C.
  & GramE\\
  
    \thickhline
     
    \end{tabular}
    \caption{Examples of unfaithful summaries per error type which correctly obtain low scores by \ours~ (top block) and incorrectly high scores (bottom block). We indicate the document sentences retrieved by \ours~ (DRS), the Model-generated Summary Sentence (MSS), and Error Type according to \cite{koh-etal-2022-far}.
    }
    \label{tab:high_low_error_types}
\end{table*}

\end{document}